\documentclass[letterpaper]{article} 
\usepackage[preprint]{aaai2027}
\usepackage[hyphens]{url}  
\usepackage{graphicx} 
\usepackage{natbib}  
\usepackage{caption} 
\usepackage{algorithm}
\usepackage{algorithmic}
\usepackage{amsfonts}
\usepackage{amsmath}
\usepackage{newfloat}
\usepackage{listings}
\usepackage{xcolor}
\usepackage{soul}
\usepackage{lineno}
\usepackage{multirow}
\usepackage[table]{xcolor}

\DeclareCaptionStyle{ruled}{labelfont=normalfont,labelsep=colon,strut=off} 
\floatstyle{ruled}
\newfloat{listing}{tb}{lst}{}
\floatname{listing}{Listing}

\usepackage{pifont}
\usepackage{makecell}
\usepackage{graphicx}
\usepackage{subcaption}
\newcommand{\cmark}{\ding{51}}
\newcommand{\xmark}{\ding{55}}

\usepackage{booktabs}

\title{SPK: Eliciting Structured Prior Knowledge for Interpretable Out-of-Distribution Detection in Real-Time Object Detection}
\author {
    Changshun Wu\textsuperscript{\rm 1,\rm 2}\corresponding,
    Weicheng He\textsuperscript{\rm 2},
    Xiaowei Huang\textsuperscript{\rm 1},
    Saddek Bensalem\textsuperscript{\rm 3}
}
\affiliations {
    \textsuperscript{\rm 1}University of Liverpool, UK\\
    \textsuperscript{\rm 2}Universit\'e Grenoble Aples, France\\
    \textsuperscript{\rm 3}CSX-AI, France\\
    Changshun.Wu@liverpool.ac.uk, Weicheng.He@univ-grenoble-alpes.fr, Xiaowei.Huang@liverpool.ac.uk, Saddek.Bensalem@csx-ai.com
}

\begin{document}

\maketitle

\begin{abstract}
Object detectors often produce over-confident predictions for objects outside their training categories, leading to so-called out-of-distribution (OoD) hallucinations. Existing approaches for detecting or mitigating such hallucinations typically either construct scoring functions directly over learned object detector representations or modify the object detector itself to suppress hallucination emergence. However, the latent priors implicitly encoded in these representations remain largely unexplored and have not been explicitly decoded for OoD detection. To uncover and exploit these latent priors, we propose Structured Prior Knowledge (SPK), a hallucination-oriented framework that explicitly elicits OoD-relevant priors from pretrained object detectors. Specifically, SPK leverages in-distribution data and hallucination-inducing samples as diagnostic supervision to elicit part-level semantic concepts underlying object detector decision-making, rather than using them merely for rejection or object detector adaptation. The elicited semantic priors are further integrated with geometric and contextual priors to form a compact five-dimensional SPK representation for OoD detection. Extensive experiments across diverse object detector architectures and multiple OoD benchmarks demonstrate that SPK achieves state-of-the-art OoD detection. Our findings reveal that pretrained object detectors already encode substantially richer latent knowledge than is typically exploited for OoD detection. More importantly, this knowledge can be explicitly elicited and organized into a compact, structured, and interpretable knowledge space for prediction reliability analysis. This suggests a promising proactive route for improving object detector reliability by explicitly uncovering and leveraging latent priors. Code and data are available at: \url{https://gricad-gitlab.univ-grenoble-alpes.fr/dnn-safety/spk}.

\end{abstract}


\section{Introduction}
\begin{figure*}[t]
    \centering
    \includegraphics[width=0.85\linewidth]{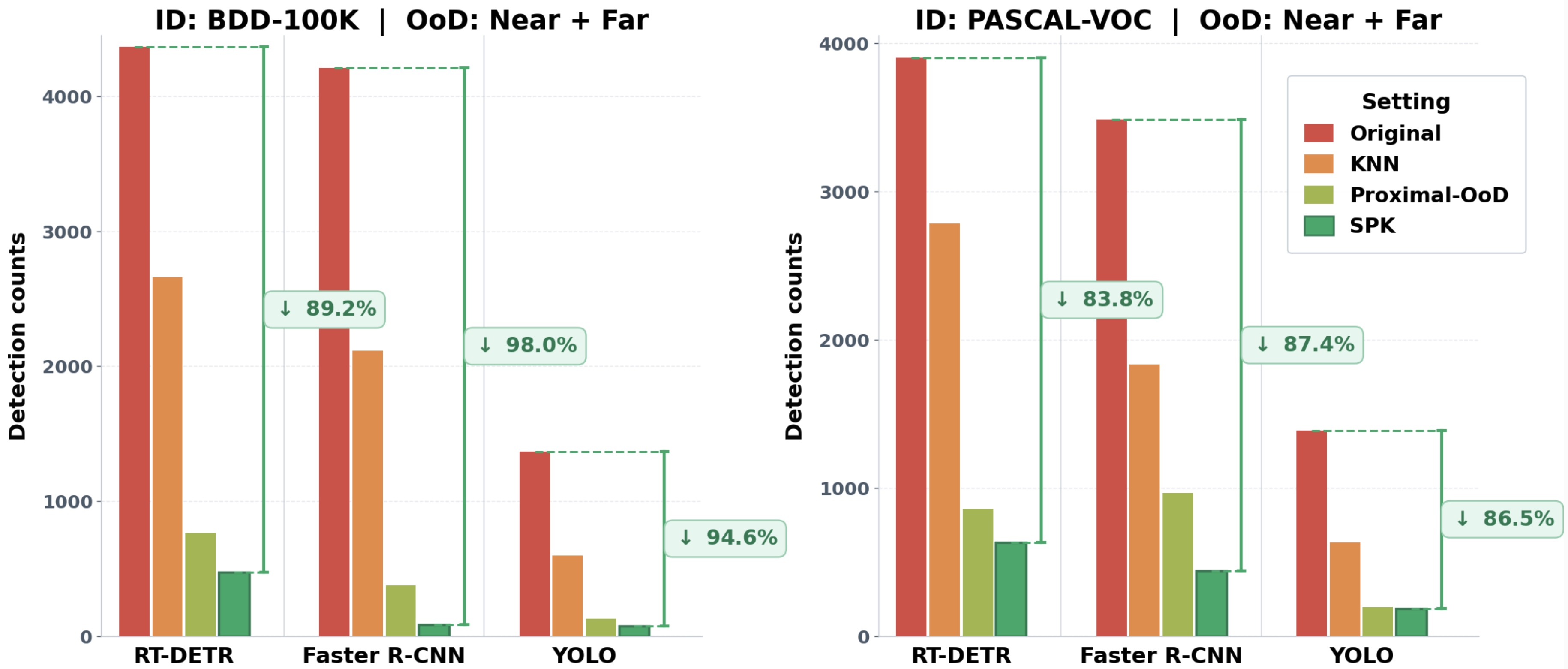}
    \caption{The proposed SPK framework, a proactive OoD hallucination mitigation framework, further reduces OoD-induced hallucinations beyond previous state-of-the-art methods~\cite{wu2026revisiting}, achieving additional improvements in challenging high-performance regimes.}
    \label{fig:overallResults}
\end{figure*}

Object detection has become a fundamental capability of modern vision systems, enabling numerous real-world applications such as autonomous driving, robotics, and intelligent perception. Despite remarkable progress, modern object detectors still suffer from a critical reliability issue: they often produce confident predictions for objects belonging to categories outside their predefined training categories. These erroneous predictions, referred to as out-of-distribution (OoD) hallucinations~\cite{he2025mitigating,wu2026revisiting}, arise when the closed-world assumption underlying detector training is violated in open-world environments. Such hallucinations may severely compromise the reliability and safety of downstream vision systems, making effective OoD hallucination mitigation an increasingly important problem.

Existing approaches mitigate OoD-induced hallucinations through two complementary routes. Reactive approaches detect hallucinations after predictions are generated by introducing external OoD detectors based on confidence estimation~\cite{Peng_2026_CVPR,Li_2026_CVPR}, uncertainty modeling~\cite{Dai_2026_CVPR}, or intermediate feature representations~\cite{du2022siren,wilson2023safe,wu2023customizable,wu2024bam,he2024box}. Proactive approaches, in contrast, seek to reduce hallucination occurrence by adapting the object detector itself, for example through lightweight fine-tuning that exploits detector-intrinsic properties~\cite{wu2026revisiting}. Although these two routes differ in their mitigation strategies, they ultimately operate on the same learned high-dimensional representation space of object detectors: existing methods either design increasingly advanced OoD scoring functions over these representations or modify the object detector to improve their discriminability.


This naturally raises a fundamental question: \emph{What latent priors have object detectors already learned within these representations that determine whether a prediction becomes a valid detection or an OoD hallucination?} We argue that answering this question provides a fundamentally different proactive perspective for OoD hallucination mitigation. Rather than designing increasingly sophisticated algorithms over opaque high-dimensional representations or modifying the object detector itself, we seek to explicitly elicit the latent priors already encoded within pretrained object detectors and organize them into a compact and interpretable representation space for OoD detection.

To answer this question, we propose Structured Prior Knowledge (SPK), a hallucination-oriented framework that explicitly elicits hallucination-relevant priors from pretrained object detectors. Our key observation is that OoD hallucinations are not arbitrary prediction failures, but are consistently associated with two representative sources. The first consists of proximal OoD objects, whose visual appearances resemble known categories and therefore induce semantic confusion. The second consists of background-only samples, where non-object regions unexpectedly activate detector objectness and produce false detections. Rather than using these samples for outlier exposure or detector adaptation, SPK treats them as diagnostic supervision for revealing the part-level semantic concepts underlying object detector decision-making. Building upon this observation, we develop an automated data construction pipeline that discovers informative proximal OoD and background-only samples while generating high-quality part-level semantic supervision. The elicited semantic priors are further integrated with geometric and contextual priors to form a compact five-dimensional SPK representation. The resulting representation consists of three semantic priors capturing part-level object concepts, a geometric prior measuring the relative spatial extent of predictions, and a contextual prior reflecting image-level contextual evidence.

Importantly, this structured representation enables stronger OoD detection performance despite its substantially lower dimensionality. As shown in Table~\ref{tab:main_fpr_table}, when applying the same downstream OoD detection algorithms, SPK consistently outperforms the original object detector representations. More importantly, as shown in Fig.~\ref{fig:overallResults}, a lightweight Isolation Forest (iForest) anomaly detector operating on the SPK representation surpasses the strongest existing mitigation strategy~\cite{wu2026revisiting}, referred to as Proximal-OoD, which combines object detector fine-tuning for suppressing hallucination emergence with KNN-based OoD detection. This advantage is consistently observed across diverse detector architectures, with clear performance margins over the existing mitigation strategy. These results highlight a new perspective for improving object detector reliability by uncovering and leveraging structured knowledge embedded within pretrained detectors. Our contributions are summarized as follows:
\begin{itemize}
    \item We propose a proactive framework for eliciting and organizing latent priors from pretrained object detectors into a structured knowledge space for OoD detection.
    \item We demonstrate the effectiveness of SPK through extensive experiments across diverse object detector architectures and multiple OoD benchmarks, achieving state-of-the-art (SoTA) performance in OoD detection while enabling lightweight and interpretable deployment.
    \item Our findings reveal that pretrained object detectors encode substantially rich latent knowledge, which can be explicitly organized into a compact, structured, and interpretable knowledge space, opening a promising new direction for improving object detector reliability.
\end{itemize}
\section{Related Work}

\paragraph{OoD Hallucination Detection and Understanding in Object Detectors} Object detectors are typically trained under a closed-world assumption and may produce over-confident predictions for objects outside their training categories, resulting in the so-called OoD hallucinations. Existing approaches can be broadly categorized into reactive and proactive paradigms. Reactive approaches formulate OoD hallucination as a post-hoc detection task (OoD detection). They detect hallucinated predictions after object detector inference by confidence estimation~\cite{Peng_2026_CVPR,Li_2026_CVPR}, uncertainty modeling~\cite{Dai_2026_CVPR}, or intermediate feature representations~\cite{du2022siren,wilson2023safe,wu2024bam}. Although effective in improving the separation between ID and OoD predictions, these methods largely treat hallucinations as prediction-level anomalies and provide limited insight into why the detector makes such decisions. Proactive approaches instead seek to understand the underlying causes of hallucination by actively diagnosing detector behavior. Recent proactive intervention methods~\cite{he2025mitigating,wu2026revisiting} reveal that conventional objectness estimation may respond to generic object-like patterns rather than ID objects, and leverage proximal OoD data to calibrate objectness toward distribution-aware prediction. In contrast, our work follows a proactive elicitation perspective. Rather than intervening in the detector's decision mechanism, we aim to uncover the latent semantic knowledge encoded in pretrained detectors by eliciting class-wise semantic prototypes responsible for their decisions. Through the SPK space, our approach provides an interpretable view of what evidence the detector observes when generating predictions and enables OoD hallucination detection based on these elicited representations.

\paragraph{Concept-based Knowledge Elicitation and Structured Priors} Understanding and extracting interpretable knowledge from deep networks has been widely studied through feature decoding and concept-based representation learning. Linear probing~\cite{alain2016linearprobes}, Network Dissection~\cite{bau2017networkdissection}, and Net2Vec~\cite{fong2018net2vec} demonstrate that intermediate representations contain human-interpretable semantic information. Concept bottleneck models further introduce explicit concept variables for improving model transparency, with recent works exploring label-free and spatial concept representations~\cite{oikarinen2023labelfreecbm,showandtell2025,ucbm2025}. Beyond interpretability, semantic concepts and structured priors have also been investigated for improving model robustness. Part-level supervision has been shown to encourage more robust object representations beyond brittle appearance cues~\cite{sitawarin2023partrobustness,li2024partglee}, while recent detector-based concept decomposition methods explore interpretable semantic evidence for unknown-object discovery~\cite{lv2026ipow}. However, existing concept-based methods are rarely designed for analyzing OoD hallucinations in object detectors. In contrast, our work leverages part-level concepts and object detector behaviors to elicit hallucination-relevant priors, integrating semantic, geometric, and contextual knowledge into a compact SPK representation for interpretable OoD detection.

\section{Problem Formulation}
We follow the standard formulation of OoD detection in object detection~\cite{du2022towards,wu2026revisiting}. Let $\mathcal{D}_{\mathrm{ID}}$ denote the ID dataset used to train an object detector, where objects belong to a predefined category set $\mathcal{C}_{\mathrm{ID}}$. Given an input image $I$, an object detector $f$ trained on $\mathcal{D}_{\mathrm{ID}}$ produces a set of predictions $\mathcal{P}=f(I)=\{p_i\}_{i=1}^{N}$, where each prediction is represented as $p_i=(\mathbf{b}_i,\hat{y}_i)$. Here, $\mathbf{b}_i$ denotes the predicted bounding box and $\hat{y}_i\in\mathcal{C}_{\mathrm{ID}}$ denotes the predicted category. For each prediction $p$, we denote its corresponding contextual image by $I_{\mathrm{ctx}}(p)$, which represents the image from which the prediction is generated.

Given an OoD dataset $\mathcal{D}_{\mathrm{OoD}}$ whose object categories are disjoint from $\mathcal{C}_{\mathrm{ID}}$, OoD detection aims to identify predictions generated on OoD inputs that should be rejected. Formally, an OoD detector $g$ is a binary decision function:
\(
g:\mathcal{P}\rightarrow\{0,1\},
\)
where an output of $1$ denotes that $p_i$ is an OoD-induced hallucination and should be rejected, while an output of $0$ denotes that $p_i$ is retained. Such OoD-induced hallucinations may originate from different sources, including objects from categories outside $\mathcal{C}_{\mathrm{ID}}$ and background regions that spuriously activate detector responses.
\section{Structured Prior Knowledge Framework}

\begin{figure*}[t]
    \centering
    \includegraphics[width=0.9\linewidth]{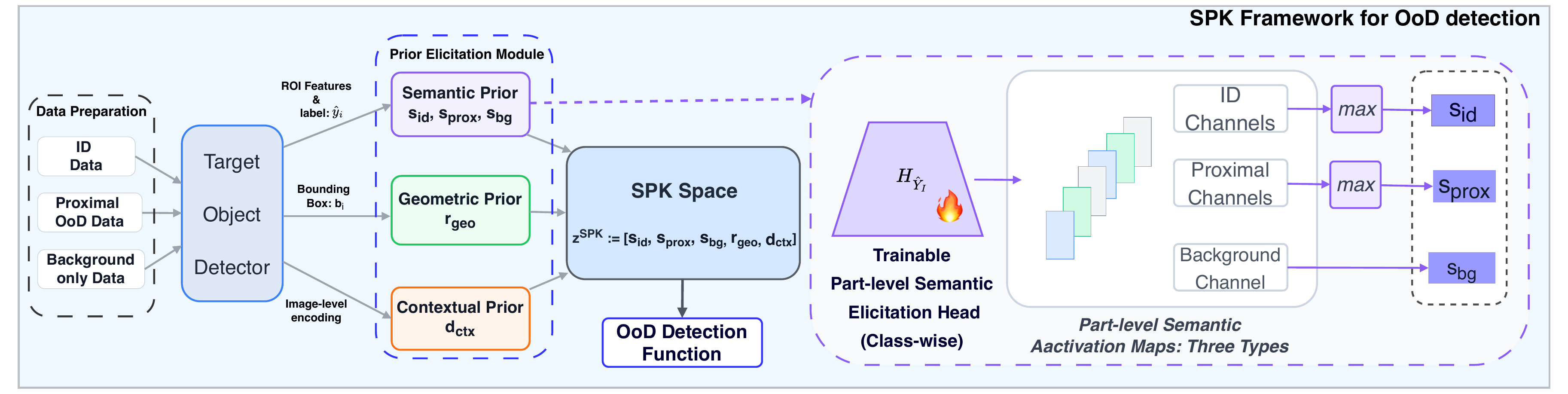}
    \caption{Overview of the proposed SPK framework. SPK elicits semantic, geometric, and contextual priors from a pretrained object detector and organizes them into a compact five-dimensional representation for OoD hallucination detection.}
    \label{fig:framework}
\end{figure*}

To answer the question of what knowledge object detectors have already acquired, we propose SPK, a proactive framework that explicitly elicits detector priors for OoD hallucination detection. Unlike existing approaches that directly operate on object detector outputs or high-dimensional feature representations, SPK aims to uncover the underlying factors that contribute to detector decisions. As illustrated in Fig.~\ref{fig:framework}, SPK elicits three complementary types of detector priors, namely semantic, geometric, and contextual priors, through hallucination-oriented supervision. These priors provide an explicit characterization of detector behavior and are organized into a compact five-dimensional SPK representation. Finally, the SPK representation is used by a lightweight OoD detector to determine whether each prediction should be retained or rejected. The following subsections describe each component of the proposed framework.

\subsection{Hallucination-oriented Data Construction}
In this section, we first detail the construction of the two types of OoD data yielding hallucinations, used along with ID data to elicit the semantic priors within our SPK framework. Then, we present the tool we built for annotating these three types data at the part-level.

\subsubsection{Hallucination-Induced Data Preparation}
We construct hallucination-inducing training data from two complementary sources: proximal OoD objects and background-only images. For each ID category, we follow the prompt design of~\cite{wu2026revisiting} and query GPT-5 to identify semantically or visually similar non-ID categories, from which annotated images are retrieved from Objects365~\cite{shao2019objects365}. We additionally collect background-only samples from the DTD dataset~\cite{Cimpoi_2014_CVPR} and filter out images containing ID objects using a pretrained object detector. We then run the target object detector on both sources and retain only the samples that induce hallucinated ID predictions. These hallucination-induced samples are used to train the semantic elicitation heads. More details on data construction and filtering are provided in Appendix~\ref{app:data_prep}.

\subsubsection{Part-Level Automated Annotation} Training the semantic elicitation module requires spatially grounded part-level semantic supervision that captures the visual evidence associated with object detector decisions. To eliminate part-level annotation as a practical bottleneck, we develop an automated annotation tool that generates localizable part concepts, grounds them within object regions, and produces the corresponding concept masks for training the semantic elicitation heads. The resulting annotations require only lightweight manual verification. Further implementation details are provided in Appendix~\ref{app:part_level_annotation}.

\subsection{Prior Elicitation}
Rather than directly constructing OoD decision functions on high-dimensional object detector representations, SPK investigates whether object detector predictions can be characterized through a small set of explicit latent priors. To this end, we elicit three complementary types of object detector priors, namely semantic, geometric, and contextual priors, each capturing a different aspect of detector behavior. Specifically, semantic priors characterize the semantic evidence associated with a prediction, geometric priors describe object-level geometric regularities, and contextual priors capture the similarity between the associated image $I_{\mathrm{ctx}}(p)$ of a prediction and ID samples in the image-level representation space. The following subsections describe each prior elicitation module.

\subsubsection{Semantic Prior Elicitation} Semantic information provides an important cue for determining whether a detector prediction corresponds to a valid object or an OoD hallucination. Inspired by recent advances in concept-based representation learning~\cite{sun2023going}, we hypothesize that object detectors also implicitly encode category-specific semantic evidence within their intermediate representations. Rather than learning new semantic knowledge, our goal is to elicit and decode this latent knowledge through part-level semantic concepts. We therefore introduce a semantic prior elicitation module that translates detector representations into structured part-level semantic responses.
Given a prediction $p_i=(\mathbf{b}_i,y_i)$, we first extract the Region-of-Interest (RoI) feature $\mathbf F_i$ corresponding to its predicted bounding box $b_i$. Instead of directly using $\mathbf F_i$ as an OoD representation, we learn a category-specific semantic elicitation head $\mathcal H_{y_i}$ to decode the latent semantic evidence encoded in the detector representation. Specifically, each semantic head predicts the activation of a set of part-level concepts associated with the prediction category.

Let \( \mathcal C_{y_i} = \{1, \ldots, N_{y_i}\} \) denote the concept vocabulary associated with prediction category $y_i$. To construct semantic supervision, we consider three complementary concept groups associated with category $y_i$: (1) part-level concepts from the ID category $y_i$, (2) concepts from proximal OoD categories related to $y_i$, and (3) background concepts obtained from background-only data. These concept groups are constructed from ID, proximal OoD, and background-only data, respectively, and represent different semantic sources that may contribute to detector predictions. For each prediction, the semantic head outputs a concept activation tensor:
\[
\mathbf A_i = \mathcal H_{y_i}(\mathbf F_i)\in[0,1]^{N_{y_i}\times H\times W},
\]
where each channel corresponds to a specific part-level concept in $\mathcal C_{y_i}$, and the spatial dimensions represent the cell-level layout within the RoI. Following part-level concept annotations, each RoI is divided into a fixed spatial grid, and the semantic head is trained to recover the corresponding cell-level concept masks. This formulation enables the detector representation to be decoded into interpretable part-level semantic responses without requiring pixel-level segmentation.

To optimize the semantic elicitation head, we employ three complementary objectives:
\[
\mathcal L_{\mathrm{SPK}}
=
\mathcal L_{\mathrm{concept}}
+
\lambda_g\mathcal L_{\mathrm{suppress}}
+
\lambda_s\mathcal L_{\mathrm{group}} .
\]

The first objective encourages accurate reconstruction of part-level semantic concepts. Specifically, we minimize a Dice loss~\cite{sudre2017generalised} between the predicted concept activation tensor $\mathbf A_i$ and the corresponding cell-level concept annotation $\mathbf M_i$:
\[
\mathcal L_{\mathrm{concept}}
=
\mathbb E_{p_i}
[
\mathrm{DiceLoss}(\mathbf A_i,\mathbf M_i)
].
\]
The Dice objective encourages each concept channel to recover the spatial distribution of its corresponding part-level concept.

The second objective prevents the semantic elicitation head from activating concepts that are not supported by the observed RoI. Given the annotated concept set of a prediction, we introduce a spurious concept suppression loss that penalizes responses from absent concepts. This objective improves the specificity of elicited semantic responses and reduces spurious concept activations. 
For prediction $p_i$, we define its absent concept set as $\mathcal C_i^-$.
For example, an RoI for a bird typically includes part-level concepts such as eyes, beak, wings, torso, and feet; however, if the feet are occluded and not visible, the ground-truth annotation for the feet concept becomes empty, and the corresponding concept channel belongs to $\mathcal C_i^-$.
Let $A_{i,c}\in[0,1]^{H\times W}$ denote the activation map corresponding to concept channel $c$, and $U_i\in\{0,1\}^{H\times W}$ the spatial union mask.
We formulate the spatial constraint as
\begin{equation}
    \mathcal L_{\mathrm{suppress}}
    =\mathbb E_{p_i}
    \left [
    \frac{1}{\left|\mathcal C_i^-\right|}
    \sum_{c\in\mathcal C_i^-}
    \frac{\left\langle
    U_i, -\log\left(1-A_{i,c}\right)
    \right\rangle
    }{
    \left\|U_i\right\|_1
    }
    \right]
\end{equation}
where $\langle\cdot,\cdot\rangle$ and $\|U_i\|_1$ denote the Frobenius inner product and the number of annotated-positive cells, respectively. The logarithm is applied element-wise. 

Finally, we introduce a group-level semantic objective to ensure that the dominant semantic response corresponds to the actual source of the underlying visual evidence. We define the set of semantic groups as $\mathcal K=\{\mathrm{id},\mathrm{prox},\mathrm{bg}\}$, where each concept channel is assigned to one group through the mapping 
\(
    \pi_{y_i}:\mathcal C_{y_i} \rightarrow \mathcal K
\).
Given the channel activation $a_i^c$, the response of each semantic group is obtained by: 
\begin{equation}
s_i^k=\max_{c\in\mathcal C_{y_i}:\pi_{y_i}(c)=k}a_i^c, \quad k\in\mathcal K 
\end{equation}

The resulting group response is optimized with a cross-entropy loss:
\[
\mathcal L_{\mathrm{group}}
=
-\mathbb E_{p_i}
\sum_{k\in\mathcal K}
q_i^k
\log
\frac{\exp(s_i^k)}
{\sum_{k'\in\mathcal K}\exp(s_i^{k'})}
\]
where $\mathbf q_i$ denotes the one-hot semantic group label of prediction $p_i$.

\subsubsection{Geometric Prior Elicitation} Object detectors inherently rely on geometric regularities when generating bounding box predictions. For example, common object detector designs exploit object scale and spatial distributions through anchor design or bounding box regression priors. Motivated by this observation, we extract geometric priors from the predicted bounding boxes to characterize whether a prediction follows the geometric patterns of known objects.
Given a prediction $p_i=(\mathbf{b_i}, \hat{y}_i)$, we compute its relative area as the geometric descriptor:
\[
r_i=\frac{\mathrm{Area}(b_i)}
{\mathrm{Area}(I_{\mathrm{ctx}}(p_i))},
\]
where $I_{\mathrm{ctx}}(p_i)$ denotes the image associated with prediction $p_i$. The relative area provides a scale-normalized measurement of the predicted object size and is invariant to image resolution. This simple geometric statistic captures the object-scale prior implicitly used by object detectors and serves as the geometric component of the proposed SPK representation.

\subsubsection{Contextual Prior Elicitation}

Besides object-level properties, detector predictions are also influenced by the visual context of the image in which they appear. We therefore introduce a contextual prior that measures the similarity between the associated image of a prediction and the ID training images.
Given a prediction $p_i$, we first extract its associated image representation: \(\mathbf{v}_i=\phi_{\mathrm{det}}(I_{\mathrm{ctx}}(p_i))\) where $\phi_{\mathrm{det}}(\cdot)$ denotes the image-level representation extracted from the detector. To focus on the priors intrinsically encoded by the detector, we directly utilize the detector representation without introducing additional external models. We construct an ID image representation bank: \( \mathcal V_{\mathrm{ID}} = \{\phi_{\mathrm{det}}(I_j)\mid I_j\in\mathcal D_{\mathrm{ID}}^{\mathrm{train}}\} \). The contextual prior is then computed by measuring the similarity between $\mathbf v_i$ and its $K$-nearest neighbors (KNN) in $\mathcal V_{\mathrm{ID}}$:
\[
d_i^{\mathrm{ctx}}
=
\operatorname{KNN}
(\mathbf v_i,\mathcal V_{\mathrm{ID}}),
\]
where $d_i^{\mathrm{ctx}}$ denotes the resulting image-level contextual score. A lower value indicates that the associated image of the prediction is more consistent with the visual characteristics observed in the ID training data.

\subsection{Structured Prior Knowledge} 

The elicited semantic, geometric, and contextual priors provide complementary views of the latent knowledge encoded by object detectors. We organize these priors into a compact SPK representation for each prediction. Specifically, the semantic module produces three group-level responses $\mathbf{s}_i=[s_i^{\mathrm{id}},s_i^{\mathrm{prox}},s_i^{\mathrm{bg}}]$, corresponding to the ID, proximal OoD, and background concept groups. Together with the geometric prior $r_i^{\mathrm{geo}}$ and contextual prior $d_i^{\mathrm{ctx}}$, the final SPK representation is defined as

\[
\mathbf z_i^{\mathrm{SPK}}
=
[
s_i^{\mathrm{id}},
s_i^{\mathrm{prox}},
s_i^{\mathrm{bg}},
r_i^{\mathrm{geo}},
d_i^{\mathrm{ctx}}
]
\in\mathbb R^5 .
\]

Each dimension of the SPK vector corresponds to an interpretable aspect of object detector knowledge, enabling OoD detection based on structured priors rather than raw high-dimensional detector features.

\subsection{SPK-based OoD Detection}

Given the structured prior representation $\mathbf z_i^{\mathrm{SPK}}$, we learn an OoD detector $g_{\mathrm{SPK}}:\mathbf z_i^{\mathrm{SPK}}\rightarrow\{0,1\}$ to determine whether a prediction should be retained or rejected. Specifically, $g_{\mathrm{SPK}}(\mathbf z_i^{\mathrm{SPK}})=1$ indicates that prediction $p_i$ is rejected as an OoD-induced hallucination, while $g_{\mathrm{SPK}}(\mathbf z_i^{\mathrm{SPK}})=0$ indicates that the prediction is retained as an ID prediction.
In practice, $g_{\mathrm{SPK}}$ can be applied with different OoD detection methods, such as KNN~\cite{sun2022out} and Isolation Forest (iForest)~\cite{liu2008isolation}. By decoupling OoD detection from raw detector features, SPK enables the final decision to be made based on explicitly elicited and structured detector priors.

\section{Experiments}
In this section, we comprehensively evaluate SPK from two perspectives: OoD detection with the underlying object detector unchanged, and OoD hallucination mitigation against approaches that modify object detector parameters. This evaluation is conducted across different object detection architectures and benchmark settings.

\subsection{Experimental Setup}

\subsubsection{OoD Detection} We follow the recently introduced calibrated evaluation protocol~\cite{wu2026revisiting}, which removes potential test contamination from OoD evaluation. Moreover, this benchmark considers three representative object detection architectures, including YOLO~\cite{wang2025yolov10}, Faster R-CNN~\cite{ren2015faster}, and RT-DETR~\cite{zhao2024detrs}, covering one-stage, two-stage, and transformer-based detectors. We evaluate on two ID detection tasks, PASCAL-VOC~\cite{everingham2010pascal} and BDD-100K~\cite{yu2020bdd100k}, with two categories of OoD test samples: Near-OoD and Far-OoD. Near-OoD samples share visual similarities with ID categories, whereas Far-OoD samples exhibit larger semantic and visual discrepancies. The baseline methods include MSP~\cite{hendrycks2017baseline}, EBO~\cite{liu2020energy}, MLS~\cite{hendrycks2022scaling}, MDS~\cite{lee2018simple}, BAM~\cite{wu2024bam}, SCALE~\cite{xu2024scaling}, KNN~\cite{sun2022out}, iForest~\cite{liu2008isolation}, and Proximal-OoD~\cite{wu2026revisiting}.

We adopt two complementary evaluation metrics. First, we report standard OoD detection metrics, including AUROC and FPR95, where FPR95 denotes the false positive rate at 95\% true positive rate of ID samples. Second, we report the reduction in the number of OoD-induced hallucinations, which enables comparison with strong OoD mitigation approaches that modify the underlying object detector. In particular, we compare with~\cite{wu2026revisiting}, which substantially improves OoD hallucination mitigation through object detector fine-tuning and represents a strong benchmark for this evaluation setting.

For comparison with OoD detection methods whose applicability to the calibrated benchmark is unclear due to architecture-specific designs or unavailable implementations, we additionally evaluate SPK under the uncalibrated benchmark adopted by these methods, despite its known annotation issues in~\cite{wu2026revisiting}. We follow the configurations in~\cite{Peng_2026_CVPR}, which considers Deformable-DETR~\cite{zhu2021deformable} on PASCAL-VOC and BDD-100K and Faster R-CNN on PASCAL-VOC, with MS-COCO~\cite{lin2014microsoft} and OpenImages~\cite{kuznetsova2020open} as OoD test sets. SPK achieves competitive performance by exploiting detector-intrinsic priors. The DINO-based variant further improves results by replacing the contextual prior with a stronger external visual representation. The complete comparison results are provided in Appendix~\ref{app:uncalibratedBenchmark}.

\subsubsection{Prior Elicitation} For semantic prior elicitation, we train a lightweight four-layer residual convolutional head that decodes each detector RoI feature to part-level semantic responses in the form of spatial concept-logit maps, one for each concept in the corresponding part-level vocabulary.  Since relevant concepts differ across object categories, we train one semantic elicitation head for each ID class. Each head is trained for up to 80 epochs with early stopping. For instance, training all heads on the 10-class BDD dataset takes about one hour on a single NVIDIA A100 GPU 40GB. Additional training details and qualitative visualizations are provided in Appendix~\ref{app:semantic_concept_learning}. To elicit contextual priors, we treat the entire image as an RoI and aggregate each multi-scale neck feature using its channel-wise spatial mean and standard deviation to form an image-level contextual embedding. We organize these embeddings into class-specific reference banks according to the object categories present in each image. More details are provided in Appendix~\ref{app:contextual_prior_learning}.

\subsection{Effectiveness of SKP}

\begin{table*}[!t]
\centering
\caption{
Comparison of OoD detection performance using FPR95 across different
detector architectures trained on PASCAL-VOC and BDD-100K.
Lower is better.
}
\label{tab:main_fpr_table}

\setlength{\tabcolsep}{3.2pt}
\renewcommand{\arraystretch}{1.08}

\resizebox{\textwidth}{!}{
\begin{tabular}{lcccc|cccc|cccc}
\toprule

\multirow{3}{*}{\textbf{Method}}
& \multicolumn{4}{c|}{\textbf{YOLO}}
& \multicolumn{4}{c|}{\textbf{Faster R-CNN}}
& \multicolumn{4}{c}{\textbf{RT-DETR}} \\

\cmidrule(lr){2-5}
\cmidrule(lr){6-9}
\cmidrule(lr){10-13}

& \multicolumn{2}{c}{\textbf{PASCAL-VOC}}
& \multicolumn{2}{c|}{\textbf{BDD-100K}}
& \multicolumn{2}{c}{\textbf{PASCAL-VOC}}
& \multicolumn{2}{c|}{\textbf{BDD-100K}}
& \multicolumn{2}{c}{\textbf{PASCAL-VOC}}
& \multicolumn{2}{c}{\textbf{BDD-100K}} \\

\cmidrule(lr){2-3}
\cmidrule(lr){4-5}
\cmidrule(lr){6-7}
\cmidrule(lr){8-9}
\cmidrule(lr){10-11}
\cmidrule(lr){12-13}

& \textbf{Near-OoD}
& \textbf{Far-OoD}
& \textbf{Near-OoD}
& \textbf{Far-OoD}
& \textbf{Near-OoD}
& \textbf{Far-OoD}
& \textbf{Near-OoD}
& \textbf{Far-OoD}
& \textbf{Near-OoD}
& \textbf{Far-OoD}
& \textbf{Near-OoD}
& \textbf{Far-OoD} \\

\midrule

MSP
& 67.48 & 67.18
& 72.93 & 74.12
& 68.36 & 78.69
& 77.46 & 74.41
& 70.44 & 67.81
& 77.46 & 74.41 \\

EBO
& 90.49 & 90.84
& 87.22 & 87.06
& 60.62 & 56.21
& 94.83 & 94.37
& 98.03 & 96.69
& 94.83 & 94.37 \\

MLS
& 89.88 & 90.08
& 86.47 & 87.06
& 59.62 & 57.89
& 92.55 & 91.08
& 92.84 & 89.75
& 92.55 & 91.08 \\

SCALE
& 80.67 & 80.92
& 77.44 & 70.59
& 92.34 & 80.70
& 86.35 & 84.98
& 81.12 & 76.92
& 86.35 & 84.98 \\

MDS
& 57.67 & 69.47
& 68.42 & 82.35
& 49.96 & 56.38
& 78.90 & 79.34
& 48.65 & 52.90
& 78.90 & 79.34 \\

BAM
& 45.36 & 43.72
& 49.63 & 52.18
& 65.44 & 42.16
& 65.73 & 61.34
& 75.61 & 65.27
& 75.48 & 68.44 \\

KNN
& 48.20 & 39.50
& 41.95 & 45.24
& 61.95 & 37.53
& 50.54 & 49.76
& 77.10 & 63.00
& 62.10 & 58.00 \\

iForest
& 70.27 & 67.82
& 60.42 & 65.23
& 75.43 & 52.38
& 63.25 & 62.78
& 79.52 & 65.92
& 68.23 & 62.30 \\
\midrule

\rowcolor[gray]{0.9}
\textbf{SPK-MDS}
& 14.99 & 17.28
& 23.35 & 6.46
& 21.03 & 23.50
& 42.42 & 42.73
& 18.43 & 17.83
& 41.77 & 16.48 \\

\rowcolor[gray]{0.9}
\textbf{SPK-BAM}
& 21.96 & 21.73
& 16.75 & 3.07
& 18.98 & 18.17
& 9.07 & 6.18
& 23.12 & 25.19
& 18.76 & 16.84 \\

\rowcolor[gray]{0.9}
\textbf{SPK-KNN}
& 19.64 & 17.99
& 13.47 & 1.17
& 15.50 & 13.69
& 4.70 & 3.09
& 18.26 & 20.43
& 16.97 & 13.74 \\

\rowcolor[gray]{0.9}
\textbf{SPK-iForest}
& \textbf{14.25} & \textbf{11.84}
& \textbf{9.86} & \textbf{0.70}
& \textbf{13.92} & \textbf{10.52}
& \textbf{2.31} & \textbf{1.52}
& \textbf{15.48} & \textbf{17.32}
& \textbf{11.42} & \textbf{9.25} \\

\bottomrule
\end{tabular}
}
\end{table*}

Rather than introducing a new OoD detector, the objective of SPK is to construct a structured prior representation on which existing OoD detection techniques can operate more effectively. We therefore evaluate SPK from two complementary perspectives: (1) whether the learned SPK representation improves the discriminability of existing OoD detection methods, and (2) whether it translates into SoTA OoD hallucination mitigation performance.

\subsubsection{A Better Representation Space for OoD Detection.} Table~\ref{tab:main_fpr_table} evaluates whether the proposed SPK representation provides a better feature space for existing OoD detection methods. We consider four representative approaches, MDS, BAM, KNN, and iForest, which can be directly applied to raw high-dimensional detector features without modifying the underlying detector. Replacing the raw detector features with the proposed SPK representation consistently yields substantial performance improvements across all detector architectures and benchmarks. Moreover, the resulting SPK-based methods outperform existing SoTA OoD detection approaches, even though the OoD detection algorithms themselves remain unchanged. These results indicate that the performance gain primarily stems from the quality of the learned representation rather than the choice of the OoD detector. This observation suggests that constructing a better representation space may be more important than designing increasingly sophisticated OoD detection algorithms. The corresponding AUROC results are provided in Appendix~\ref{app:moreResults}, leading to the same conclusions.

\begin{table}[t]
\centering
\caption{OoD detection counts (Near-OoD/Far-OoD) across different detector architectures. Lower is better.}
\label{tab:multi_arch_ood_counts_updated}
\resizebox{0.9\columnwidth}{!}{
\begin{tabular}{llcc}
\toprule
\textbf{Model}
& \textbf{Method}
& \textbf{VOC (N/F)}
& \textbf{BDD (N/F)} \\
\midrule

\multirow{3}{*}{YOLO}
& Original
& 946 / 440
& 701 / 666 \\
& Proximal-OoD
& \textbf{134} / 60
& 80 / 47 \\
& \cellcolor[gray]{0.9}\textbf{SPK}
& \cellcolor[gray]{0.9} 135 / \cellcolor[gray]{0.9}\textbf{52}
& \cellcolor[gray]{0.9}\textbf{69} / \cellcolor[gray]{0.9}\textbf{5} \\
\midrule

\multirow{3}{*}{\makecell{Faster\\R-CNN}}
& Original
& 2150 / 1335
& 2576 / 1634 \\
& Proximal-OoD
& 710 / 253
& 207 / 167 \\
& \cellcolor[gray]{0.9}\textbf{SPK}
& \cellcolor[gray]{0.9}\textbf{299} / \textbf{140}
& \cellcolor[gray]{0.9}\textbf{60} / \textbf{25} \\
\midrule

\multirow{3}{*}{RT-DETR}
& Original
& 2311 / 1589
& 3145 / 1220 \\
& Proximal-OoD
& 386 / 470
& 525 / 240 \\
& \cellcolor[gray]{0.9}\textbf{SPK}
& \cellcolor[gray]{0.9}\textbf{358} / \cellcolor[gray]{0.9}\textbf{275}
& \cellcolor[gray]{0.9}\textbf{359} / \cellcolor[gray]{0.9}\textbf{113} \\
\bottomrule
\end{tabular}
}
\end{table}

Beyond standard OoD detection metrics, we further evaluate the practical effectiveness of SPK by measuring the reduction in the number of OoD-induced hallucinations. As shown in Table~\ref{tab:multi_arch_ood_counts_updated}, SPK consistently removes substantially more hallucinations than the recent ProximalOoD-based fine-tuning approach~\cite{wu2026revisiting} across three representative detector architectures. Notably, SPK achieves these improvements without modifying or fine-tuning the underlying object detector, demonstrating that explicitly leveraging the object detector's latent prior knowledge alone is sufficient to achieve highly effective OoD mitigation.

\subsection{Ablation on Semantic Prior Elicitation} 

The semantic prior is the core component of the proposed SPK framework. To investigate the contribution of each learning objective, we conduct an ablation study by progressively removing individual loss terms while keeping all other components unchanged. The results are summarized in Table~\ref{tab:loss_ablation}. Removing any of the three losses consistently degrades OoD detection performance, demonstrating that they play complementary roles during semantic prior elicitation. Specifically, the Dice loss provides accurate supervision for part-level concept activation, the group discrimination loss encourages concept responses to concentrate on the correct semantic group, and the spatial consistency loss further regularizes the spatial activation patterns within each concept channel. The best performance is achieved only when all three objectives are jointly optimized, confirming that accurate semantic prior elicitation requires both semantic discrimination and spatially consistent concept representations. Additional quantitative results on Faster R-CNN and RT-DETR are provided in the Appendix~\ref{app:moreResults}, showing consistent trends across different detector architectures.

\begin{table}[t]
\centering
\caption{Ablation study of the SPK loss components on YOLO trained on
PASCAL-VOC and BDD-100K. Results are reported as Near-OoD / Far-OoD
FPR95. The average is computed over all four results. Lower is better.}
\label{tab:loss_ablation}

\setlength{\tabcolsep}{3.2pt}
\renewcommand{\arraystretch}{1.08}

\resizebox{0.9\linewidth}{!}{
\begin{tabular}{ccc|ccc}
\toprule
$\mathcal{L}_{\mathrm{dice}}$
& $\mathcal{L}_{\mathrm{suppress}}$
& $\mathcal{L}_{\mathrm{group}}$
& \textbf{VOC}
& \textbf{BDD}
& \textbf{Average} \\[-1pt]

& & &
{\scriptsize Near / Far}
& {\scriptsize Near / Far}
& {\scriptsize FPR95 $\downarrow$} \\
\midrule

\xmark & \cmark & \cmark
& 25.80 / 21.30
& 17.85 / 18.26
& 20.80 \\

\cmark & \cmark & \xmark
& 22.10 / 16.40
& 15.29 / 15.97
& 17.44 \\

\cmark & \xmark & \cmark
& 20.50 / 15.80
& 14.18 / 12.93
& 15.85 \\

\cmark & \cmark & \cmark
& \textbf{14.25 / 11.84}
& \textbf{9.86 / 0.70}
& \textbf{9.16} \\

\bottomrule
\end{tabular}
}
\end{table}

\subsection{Ablation on Prior Components}

We next investigate the contribution of each prior component in the proposed SPK representation. Starting from semantic priors alone, we progressively incorporate geometric and contextual priors while keeping the OoD detector unchanged. The results are reported in Table~\ref{tab:ablation_priors}. Semantic priors alone already provide strong OoD discrimination, confirming that part-level concepts capture rich latent knowledge encoded by object detectors. Incorporating geometric priors further improves the performance, suggesting that object-scale information provides complementary structural cues beyond semantic concepts. Adding contextual priors brings further improvements, particularly on the more complex BDD-100K benchmark, where image-level similarity provides additional cues for distinguishing OoD-induced hallucinations from valid ID predictions. In contrast, the improvement on PASCAL-VOC is relatively limited, potentially because its test images share similar visual characteristics with the ID distribution, reducing the additional discriminative benefit provided by image-level context. Overall, these results demonstrate that semantic, geometric, and contextual priors capture complementary aspects of detector knowledge, and their combination provides a more comprehensive representation of the latent priors encoded by object detectors. Additional quantitative results on Faster R-CNN and RT-DETR are provided in the Appendix~\ref{app:moreResults}, showing consistent trends across different detector architectures.

\begin{table}[h]
\centering
\caption{Ablation study of different prior components on YOLO trained on
PASCAL-VOC and BDD-100K. Each dataset column reports Near-OoD / Far-OoD
FPR95. The average is computed over all four results. Lower is better.}
\label{tab:ablation_priors}

\setlength{\tabcolsep}{4.5pt}
\renewcommand{\arraystretch}{1.30}

\resizebox{\columnwidth}{!}{
\begin{tabular}{l|ccc}
\toprule
\textbf{Prior components}
& \textbf{VOC}
& \textbf{BDD}
& \textbf{Average} \\[-1pt]

& {\scriptsize Near / Far}
& {\scriptsize Near / Far}
& {\scriptsize FPR95 $\downarrow$} \\
\midrule

{\scriptsize\bfseries Semantic}
& 15.43 / 13.28
& 30.37 / 25.58
& 21.17 \\

{\scriptsize\bfseries Semantic + Geometric}
& \textbf{13.23 / 11.50}
& 20.88 / 3.84
& 12.36 \\

{\scriptsize\bfseries Semantic + Geometric + Contextual}
& 14.25 / 11.84
& \textbf{9.86 / 0.70}
& \textbf{9.16} \\

\bottomrule
\end{tabular}
}
\end{table}

\section{Conclusion}
In this work, we propose Structured Prior Knowledge (SPK), a proactive framework that explicitly elicits latent priors from pretrained object detectors for OoD detection. By organizing these priors into a compact and interpretable representation, SPK enables effective OoD detection and hallucination mitigation without modifying the underlying detector. Extensive experiments across diverse object detector architectures and benchmark settings demonstrate the effectiveness and generality of the proposed framework. Beyond the framework itself, our results reveal that pretrained object detectors encode substantially richer knowledge than is typically exploited for OoD detection. Rather than learning new representations from raw high-dimensional detector features, SPK shows that these latent priors can be explicitly elicited into an interpretable knowledge space, where each dimension corresponds to a semantically meaningful quantity. We believe this study establishes prior elicitation as a promising proactive direction for OoD detection and suggests that exposing latent model knowledge may provide a general strategy for improving the reliability and interpretability of foundation models.
While this work focuses on two representative sources of OoD-induced hallucinations, namely proximal OoD objects and background-induced false predictions, the proposed prior elicitation framework can naturally be extended to additional hallucination sources and broader reliability challenges. Future work will investigate how to uncover richer latent priors and how to better leverage the elicited knowledge to prevent OoD hallucination generation.

\section*{Acknowledgments}
Funded by the European Union. Views and opinions expressed are however those of the author(s) only and do not necessarily reflect those of the European Union or the European Health and Digital Executive Agency (HADEA). Neither the European Union nor the granting authority can be held responsible for them. RobustifAI project, ID 101212818.

\bibliography{aaai2027}


\clearpage
\appendix
\refstepcounter{section}
\section*{Appendix \thesection}
\refstepcounter{subsection}
\subsection*{\thesubsection\quad Data Preparation}
\label{app:data_prep}

\subsubsection{Proximal mining} \textit{Proximal OoD} objects are one of the main sources of detector hallucinations because they share semantic or visual characteristics with ID categories. 
To curate such data, we follow the prompt design introduced in~\cite{wu2026revisiting} and query GPT-5 to generate a set of proximal categories set for each ID category. 
The proposed proximal categories should be disjoint from the ID categories while being semantically or visually similar to the corresponding ID category. 
For each ID category, we use the obtained proximal category set to retrieve 1,000 annotated images from Objects365~\cite{shao2019objects365}, a large-scale object detection dataset covering diverse object categories. 
As our evaluation benchmarks are constructed from Open Images V7~\cite{benenson2019large}, using Objects365 as the source of training data helps avoid any data leakage issue.
We run the target object detector on these images and retain images that induce hallucinated ID predictions.

\subsubsection{Background mining} OoD hallucinations in object detection may also occur when object detectors encounter backgrounds with textures or local cues resembling those of ID objects. 
To capture this failure mode, we collect \textit{background-only samples} from the Describable Textures Dataset (DTD)~\cite{Cimpoi_2014_CVPR}. 
DTD is an image-level texture recognition dataset containing diverse real-world texture patterns, making it a good source for probing background-induced detector hallucinations.
Since DTD is an image classification dataset without object-level annotations and may contain ID objects, we first  employ YOLOE-11-L~\cite{wang2025yoloe} to filter out images containing ID objects.

\subsubsection{Part-Level Automated Annotation} 
\label{app:part_level_annotation}

\begin{figure*}[t]
    \centering
    \includegraphics[width=\textwidth]{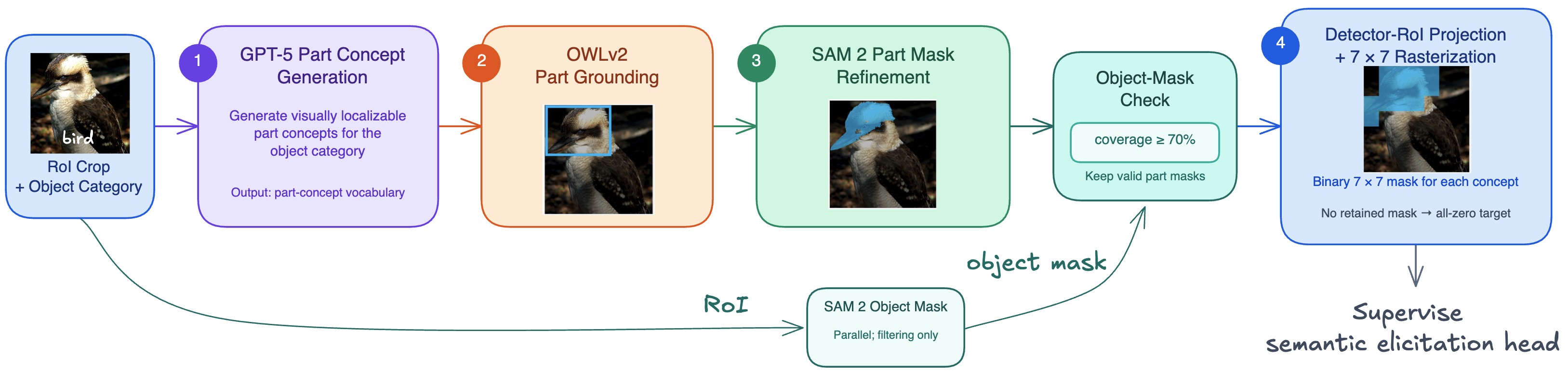}
    \caption{\textbf{Automated part-level annotation pipeline.} Given an RoI crop and its object category, OWLv2 grounds the corresponding GPT-5-generated concept vocabulary into part bounding boxes. Each box prompts SAM\,2 to produce a refined pixel-level part mask. Masks satisfying the object-mask coverage threshold are projected into detector-RoI coordinates and rasterized as binary $7 \times 7$ targets. Concepts without a retained mask receive an all-zero target.}
    \label{fig:auto_annotate}
\end{figure*}

To train the semantic elicitation head, we require part-level semantic annotations that capture the visual evidence associated with object detector predictions. Since manually annotating such concepts is labor-intensive and difficult to scale, we develop an automated part-level annotation tool, illustrated in Fig.~\ref{fig:auto_annotate}, to efficiently generate the required supervision. Given an object category, our tool first queries GPT-5 following the prompt design of~\cite{oikarinen2023labelfreecbm} to generate a compact vocabulary of visually localizable part-level concepts. For example, for the category \textit{bird}, the generated concepts include \textit{beak}, \textit{eye}, \textit{wing}, \textit{torso}, and \textit{feet}. The tool then grounds each concept within the corresponding Region of Interest (RoI) using OWLv2~\cite{minderer2023scaling}. Each grounded part is subsequently refined into a pixel-level segmentation mask by SAM\,2~\cite{ravi2025sam}. To suppress background-induced annotations, the generated part mask is retained only if at least 70\% of its pixels overlap with the corresponding object mask predicted by SAM\,2. Finally, each retained part mask is projected into object detector-RoI coordinates and converted into the concept-level supervision used to train the semantic elicitation head. Concepts without valid annotations are assigned empty supervision.

We evaluate the quality of the proposed annotation tool in Table~\ref{tab:part_annotation_quality}. Our tool consistently outperforms existing automatic part annotation methods, achieving the highest overall mean Intersection over Union (mIoU) and Recall@0.5. Compared with the strongest baseline, VLPart~\cite{sun2023going}, it improves mIoU and Recall@0.5 by 3.8 and 1.6 percentage points, respectively. The improvement is consistent across object categories, with our tool achieving the best Concept-mIoU on six of the seven evaluated classes. These results demonstrate that the proposed tool can automatically generate accurate part-level semantic annotations, making semantic elicitation practical without requiring extensive manual annotation.

\begin{figure*}[!h]
  \centering
  \includegraphics[width=0.9\linewidth]{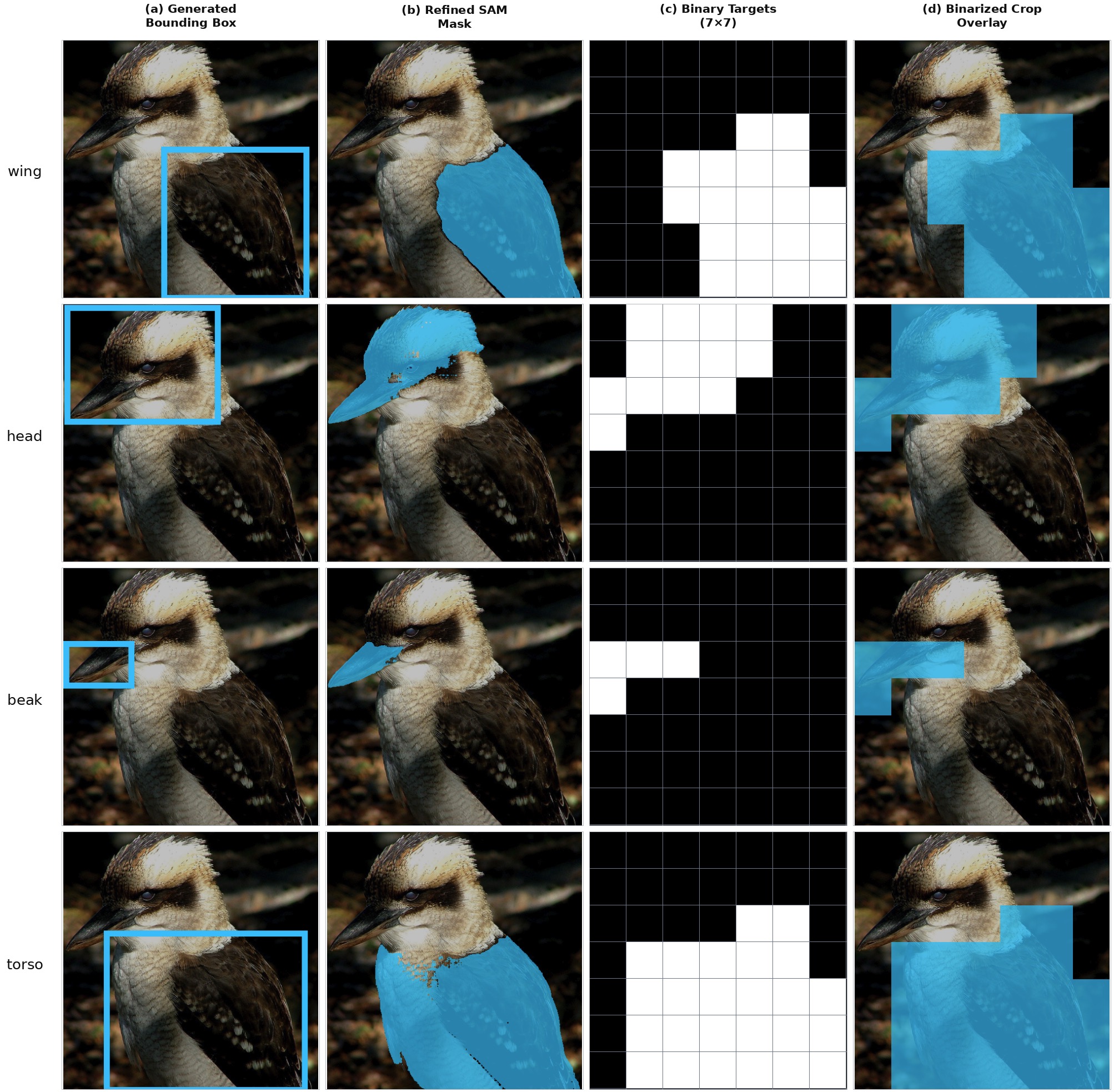}
  \caption{\textbf{Part-Level Concept Annotation Example.}
  We use a \texttt{bird} RoI crop to illustrate the step-by-step annotation process for the concepts \textit{wing}, \textit{head}, \textit{beak}, and \textit{torso}. Step 1: OWLv2 grounds each concept within the detector RoI and generates a bounding-box proposal (a). Step 2: SAM\,2 refines each proposal into a pixel-level part mask, which is retained only if at least 70\% of its pixels overlap with the corresponding SAM\,2 object mask (b). Step 3: Each retained mask is projected into detector-RoI coordinates and converted into a binary $7 \times 7$ supervision target (c). Step 4: The binary target is overlaid on the RoI crop for visualization (d).}
  \label{fig:auto_annotate}
\end{figure*}

\begin{table*}[t]
\centering
\small
\caption{\textbf{Quality of the automated part-level annotations.} We report the
overall concept-mIoU, Recall@0.5, number of covered concepts, and per-class
concept-mIoU. All values except concept coverage are percentages. Best
results are shown in bold.}
\label{tab:part_annotation_quality}
\setlength{\tabcolsep}{4.2pt}
\renewcommand{\arraystretch}{1.08}
\resizebox{\textwidth}{!}{%
\begin{tabular}{@{}lccc|ccccccc@{}}
\toprule
\multirow{2}{*}{\textbf{Method}}
& \multicolumn{3}{c|}{\textbf{Overall}}
& \multicolumn{7}{c}{\textbf{Per-class concept-mIoU}} \\
\cmidrule(lr){2-4}
\cmidrule(lr){5-11}
& \textbf{concept-mIoU}
& \textbf{Recall@0.5}
& \textbf{\# Concepts}
& \textbf{bird}
& \textbf{bus}
& \textbf{car}
& \textbf{cat}
& \textbf{cow}
& \textbf{dog}
& \textbf{horse} \\
\midrule
Ours
& \textbf{40.8} & \textbf{38.4} & \textbf{25}
& \textbf{46.5} & \textbf{37.9} & \textbf{43.9} & \textbf{40.4}
& \textbf{37.4} & 41.8 & \textbf{38.2} \\
VLPart~\cite{sun2023going}
& 37.0 & 36.8 & \textbf{25}
& 35.6 & 32.8 & 32.5 & 38.1 & 36.8 & \textbf{44.1} & 36.5 \\
Grounded SAM~\cite{ren2024grounded}
& 32.3 & 30.7 & \textbf{25}
& 32.1 & 31.7 & 31.3 & 33.0 & 29.5 & 38.3 & 29.4 \\
\bottomrule
\end{tabular}%
}
\end{table*}
\refstepcounter{subsection}
\subsection*{\thesubsection\quad Implementation Details of Prior Elicitation}

\subsubsection{Multi-Scale RoI Feature Extraction} An important requirement of our semantic elicitation head is to reconstruct spatial concept evidence inside each detected proposal, which makes the choice of proposal-level representation crucial. Let $p_i=(\mathbf{b}_i, y_i)$ denote a detector prediction and let $\mathbf{F}_i$ denote the RoI feature associated with its predicted bounding box $\mathbf{b}_i$. 
For YOLO and RT-DETR, we extract features from three spatial feature levels, denoted by $\{\mathbf{E}^{\ell}\}_{\ell=1}^{L}$, where \[
\mathbf{E}^{\ell}
\in
\mathbb{R}^{C_{\ell}\times H_{\ell}\times W_{\ell}}
\]
Specifically, we use the detection-neck outputs for YOLO and the multi-scale
hybrid-encoder outputs for RT-DETR. 
For the latter, we use the hybrid-encoder
features instead of the final decoder queries because decoder queries do not
define a canonical spatial grid within the predicted bounding
box~\cite{zhao2024detrs}.
For each prediction, RoIAlign is independently applied to every feature level:
\begin{equation}
    \mathbf{F}_{i}^{\ell} = \operatorname{RoIAlign}
    \left(
        \mathbf{E}^{\ell},
        \mathbf{b}_i;
        7\times7
    \right)
    \in 
    \mathbb{R}^{C_{\ell}\times7\times7}.
\end{equation}

The multi-scale RoI features are then concatenated along the channel dimension: 
\begin{equation}
    \mathbf{F}_i = \operatorname{Concat}_{\ell=1}^{L}
    \left(
        \mathbf{F}_{i}^{\ell}
    \right)
    \in 
    \mathbb{R}^{D\times7\times7}, ~
    D=\sum_{\ell=1}^{L}C_{\ell}.
\end{equation}

This produces an RoI representation that combines fine-grained spatial
evidence with coarser semantic context. 
For Faster R-CNN, we directly use the RoI feature produced by the Detectron2 \texttt{box\_pooler}. 
Following the standard FPN assignment strategy, each predicted box is assigned to an appropriate pyramid level and pooled into a $256\times7\times7$ feature map.
Table~\ref{tab:roi_feature_layers} summarizes the feature sources, channels, and resulting RoI feature dimensions for all detector architectures.

\begin{table}[!h]
\centering
\small
\caption{
\textbf{RoI feature sources and dimensions for each detector.}
YOLO and RT-DETR concatenate RoI-aligned features from three feature scales,
whereas Faster~R-CNN uses the Detectron2 \texttt{box\_pooler}.
}
\label{tab:roi_feature_layers}
\resizebox{\columnwidth}{!}{%
\begin{tabular}{@{}llll@{}}
\toprule
\textbf{Detector}
& \textbf{Feature source}
& \textbf{Channels}
& \textbf{RoI feature $\mathbf{F}_i$} \\
\midrule
YOLO
& Detect neck
& $128+256+512$
& $896\times7\times7$ \\
RT-DETR
& Hybrid-encoder neck
& $256+256+256$
& $768\times7\times7$ \\
Faster R-CNN
& ResNet-FPN
& $256$
& $256\times7\times7$ \\
\bottomrule
\end{tabular}%
}
\end{table}

\subsubsection{Part-Level Semantic Concept Learning.}
\label{app:semantic_concept_learning}
For YOLOv10, the multi-scale RoI feature map has dimension
$\mathbf{F}_i \in \mathbb{R}^{896 \times 7 \times 7}$. The resulting representation is
processed by a lightweight semantic elicitation head comprising a
$1 \times 1$ projection layer, two residual convolutional blocks, and a final
$1 \times 1$ prediction layer. The projection layer reduces the channel
dimension from 896 to 256, while each residual block applies a
$3 \times 3$ convolution followed by GroupNorm, GELU, and Dropout2d. The head
produces a spatial concept-logit tensor
$z \in \mathbb{R}^{C \times 7 \times 7}$,
where $C$ denotes the size of the corresponding part-level concept vocabulary. Because part vocabularies are inherently class-specific, e.g., \emph{wing}, \emph{tail}, and \emph{beak} are relevant to birds, whereas \emph{body} and \emph{cap} are relevant to bottles, we train an independent semantic elicitation head for each ID category. To obtain an RoI-level activation for each concept, we aggregate its spatial logit map using log-sum-exp.
The resulting vector
captures the activation of category-specific semantic concepts within the RoI and is subsequently used as the semantic prior in SPK. Each head is trained for up to 80 epochs with early stopping. Training all heads takes approximately one hour on a single NVIDIA A100 GPU.
Additional architectural and training hyperparameters of the semantic elicitation head are summarized in Table~\ref{tab:experiment_hyperparameters}.

\begin{table*}[h]
\centering
\caption{Hyperparameters of the Semantic Elicitation Head.}
\label{tab:experiment_hyperparameters}
\renewcommand{\arraystretch}{1.15}
\begin{tabular}{p{0.25\textwidth} l l}
\hline
\textbf{Semantic Elicitation Head} & \textbf{Hyperparameter} & \textbf{Value} \\
\hline
\multirow{10}{0.25\textwidth}{\centering Architecture}
& Hidden channels      & \textbf{256} \\
& Dropout              & \textbf{0.1} \\
& Normalization        & \textbf{GroupNorm} \\
& Activation           & \textbf{GELU} \\
& Residual blocks      & \textbf{2} \\
& Output head          & \textbf{$1 \times 1$ conv $\rightarrow$ num\_concepts} \\
& Inference activation & \textbf{Sigmoid} \\
& RoI spatial size     & \textbf{$7 \times 7$} \\
& Inference pooling    & \textbf{LogSumExp, $\tau = 0.5$} \\
\hline
\multirow{11}{0.25\textwidth}{\centering Training}
& Training epochs         & \textbf{80} \\
& Batch size              & \textbf{2000} \\
& Optimizer               & \textbf{AdamW} \\
& Learning rate           & \textbf{$2 \times 10^{-4}$} \\
& Weight decay            & \textbf{$5 \times 10^{-4}$} \\
& Random seed             & \textbf{42} \\
& Validation split        & \textbf{10\%} \\
& Training sampler        & \textbf{WeightedRandomSampler} \\
& Suppress loss weight    & \textbf{0.25} \\
& Group loss weight       & \textbf{0.75} \\
& Early-stopping patience & \textbf{10 epochs} \\
\hline
\end{tabular}
\end{table*}

\subsubsection{Contextual Prior Learning.}
\label{app:contextual_prior_learning}
 
To elicit contextual priors, we treat the entire image as an RoI and construct an image-level representation from the detector's multi-scale neck features. 
For each feature scale, we compute the channel-wise mean and standard deviation over all spatial cells and concatenate these statistics across scales to form an embedding vector $\mathbf v$, which is subsequently $\ell_2$-normalized. 
The detector-specific feature sources and resulting embedding dimensions are summarized in Table~\ref{tab:image_embedding_layers}. 
Using ID training images, we organize the normalized embeddings into class-specific reference banks $\{\mathcal V_{\mathrm{ID}}^{y}\}_{y\in\mathcal Y}$.  
At retrieval time, neighbors are ranked by cosine similarity. We retain its $k$ nearest references and define the contextual distance as
\begin{equation}
    d_{\mathrm{ctx}}(\mathbf{v})
    =
    1-
    \frac{1}{k}
    \sum_{\mathbf{r}\in\mathcal{N}_{k}(\mathbf{v},\mathcal V_{\mathrm{ID}}^{\hat y}))}
    \operatorname{cos}\!\left(\mathbf{v},\mathbf{r}\right),
    \label{eq:contextual_distance}
\end{equation}
where $\hat{y}$ is the predicted class, and
$\mathcal{N}_{k}(\mathbf{v},\mathcal V_{\mathrm{ID}}^{\hat y})$
contains the $k$ nearest reference embeddings from the class-specific reference bank $\mathcal V_{\mathrm{ID}}^{\hat y}$. We set $k=5$ in all experiments. A small contextual distance indicates that the query occurs in a context similar to those observed for the predicted class during ID training, whereas a large distance indicates an atypical context. We use this distance as the contextual prior in SPK.

\begin{table*}[!h]
\centering
\small
\caption{
\textbf{Image-level embedding sources and dimensions for each detector.}
We globally pool each selected feature map by its spatial mean and standard
deviation, then concatenate the resulting statistics. YOLO and RT-DETR use
their deepest selected backbone stage, whereas Faster~R-CNN aggregates all
four ResNet-FPN levels.
}
\label{tab:image_embedding_layers}
\resizebox{0.65\linewidth}{!}{%
\begin{tabular}{@{}llll@{}}
\toprule
\textbf{Detector}
& \textbf{Feature source}
& \textbf{Channels}
& \textbf{Embedding dimension} \\
\midrule
YOLO
& Backbone L6 (stride $16$)
& $256$
& $\mathrm{mean}+\mathrm{std}: 512$ \\
RT-DETR
& HGBlock L9 backbone (stride $32$)
& $2048$
& $\mathrm{mean}+\mathrm{std}: 4096$ \\
Faster R-CNN
& ResNet-FPN P2--P5
& $4\times256$
& $\mathrm{mean}+\mathrm{std}: 2048$ \\
\bottomrule
\end{tabular}%
}
\end{table*}

\refstepcounter{subsection}
\subsection*{\thesubsection\quad Semantic Prior Learning Analysis}

\begin{figure}[h]
    \centering

  \begin{subfigure}[t]{0.485\linewidth}
    \centering
    \includegraphics[width=\linewidth]{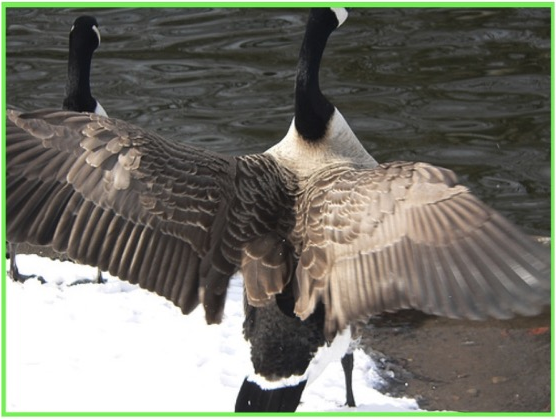}
    \caption{Input + ROI}
    \label{fig:qualitative_input_roi}
  \end{subfigure}
  \hfill
  \begin{subfigure}[t]{0.485\linewidth}
    \centering
    \includegraphics[width=\linewidth]{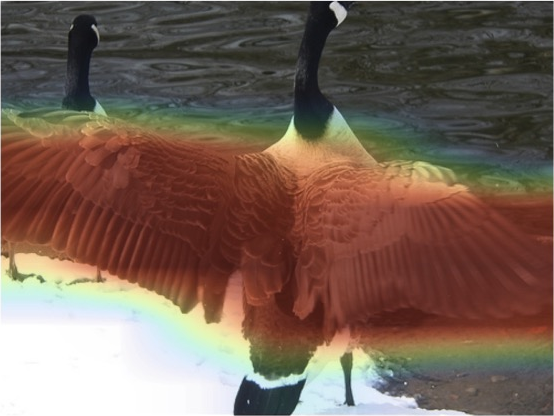}
    \caption{Prediction: wing}
    \label{fig:qualitative_wing}
  \end{subfigure}

  \vspace{0.5em}

  \begin{subfigure}[t]{0.485\linewidth}
    \centering
    \includegraphics[width=\linewidth]{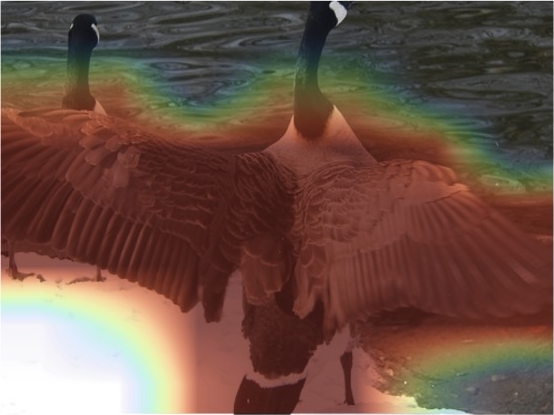}
    \caption{Prediction: torso}
    \label{fig:qualitative_torso}
  \end{subfigure}
  \hfill
  \begin{subfigure}[t]{0.485\linewidth}
    \centering
    \includegraphics[width=\linewidth]{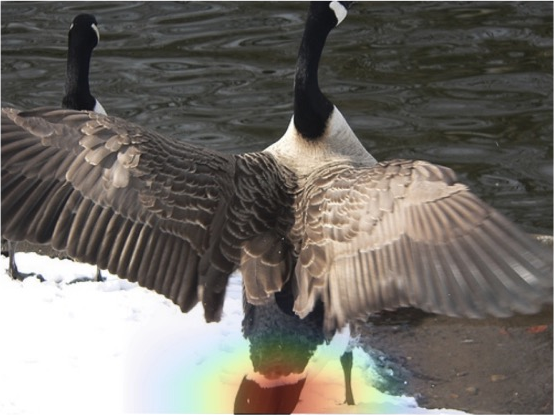}
    \caption{Prediction: foot}
    \label{fig:qualitative_foot}
  \end{subfigure}
    \caption{
        \textbf{Qualitative visualization of part-level semantic responses.} The learned concept maps are well aligned with the corresponding regions, demonstrating that the semantic elicitation head successfully decodes spatially grounded semantic evidence from detector RoI features.}
    \label{fig:concept_visualization}
\end{figure}

We evaluate the learned semantic elicitation heads from two complementary perspectives: (1) the quality of the elicited part-level semantic responses and (2) the discriminative capability of the learned semantic groups. Fig.~\ref{fig:concept_visualization} visualizes the concept activation maps predicted by the bird-specific semantic elicitation head. The activations corresponding to concepts such as \emph{wing}, \emph{torso}, and \emph{foot} are well aligned with their anatomical regions, indicating that the semantic head successfully decodes spatially grounded part-level semantic evidence from the detector RoI representation.

We further evaluate whether the learned semantic responses are correctly summarized into the three semantic groups introduced by the group objective. Table~\ref{tab:semantic_head_softmax} reports the group prediction accuracy on ID, proximal OoD, and background samples. The semantic head achieves high group prediction accuracy on the training data, indicating that the learned concept responses are effectively organized according to their underlying semantic source. Although the accuracy decreases on unseen validation samples, the overall trend confirms that the proposed group objective successfully captures the intended semantic grouping.

\begin{table}[t]
\centering
\caption{\textbf{Semantic-head group-classification accuracy.}
Computed by applying $\arg\max$ to concept activations for YOLO on PASCAL-VOC.}
\label{tab:semantic_head_softmax}
\begin{tabular}{lc}
    \toprule
    \cellcolor{white}\textbf{Data split} & \textbf{Accuracy (\%)} \\
    \midrule
    ID training set    & 96.5 \\
    Proximal OoD       & 81.0 \\
    Background            & 77.0 \\
    \bottomrule
\end{tabular}
\end{table}

\begin{figure}[!h]
    \centering
    \includegraphics[width=\linewidth]
        {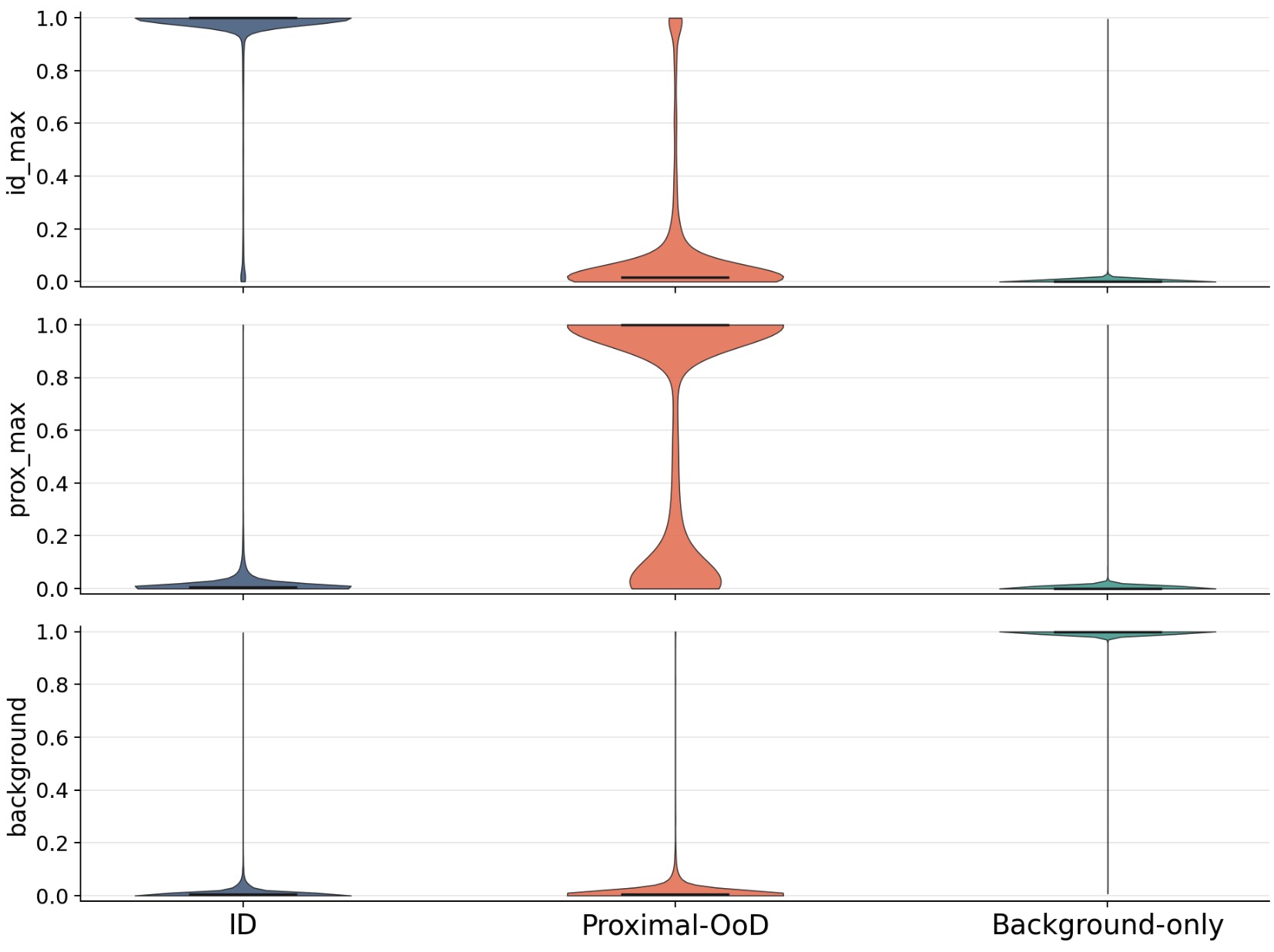}
    \caption{ \textbf{Distributions of the learned semantic group responses.} Samples from different data sources predominantly activate their corresponding semantic groups, validating the effectiveness of the proposed group objective.}
    \label{fig:semantic-head-activations}
\end{figure}

Fig.~\ref{fig:semantic-head-activations} further analyzes the learned semantic group responses on the training data. ID samples are dominated by the ID semantic group, while the responses of the proximal and background groups remain largely suppressed. Proximal OoD samples predominantly activate the proximal semantic group, with a small fraction also exhibiting strong ID responses due to their high semantic similarity to ID objects. Likewise, background samples are characterized by dominant background responses and consistently low activations for the ID and proximal groups. These results demonstrate that the learned semantic groups successfully capture the underlying source of the visual evidence and provide meaningful semantic priors for constructing the SPK representation.

\begin{figure*}[t]
    \centering

    \begin{subfigure}{0.95\textwidth}
        \centering
        \includegraphics[width=\textwidth]{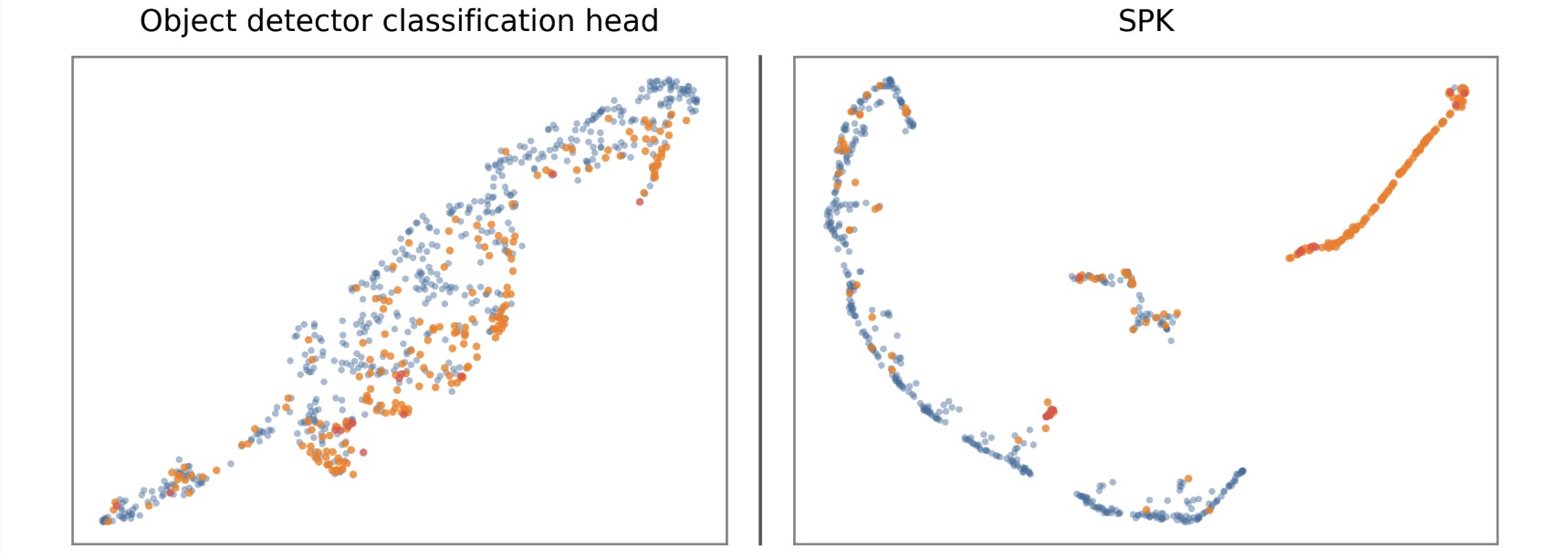}
        \caption{\emph{dog}}
    \end{subfigure}

    \vspace{0.5em}

    \begin{subfigure}{0.95\textwidth}
        \centering
        \includegraphics[width=\textwidth]{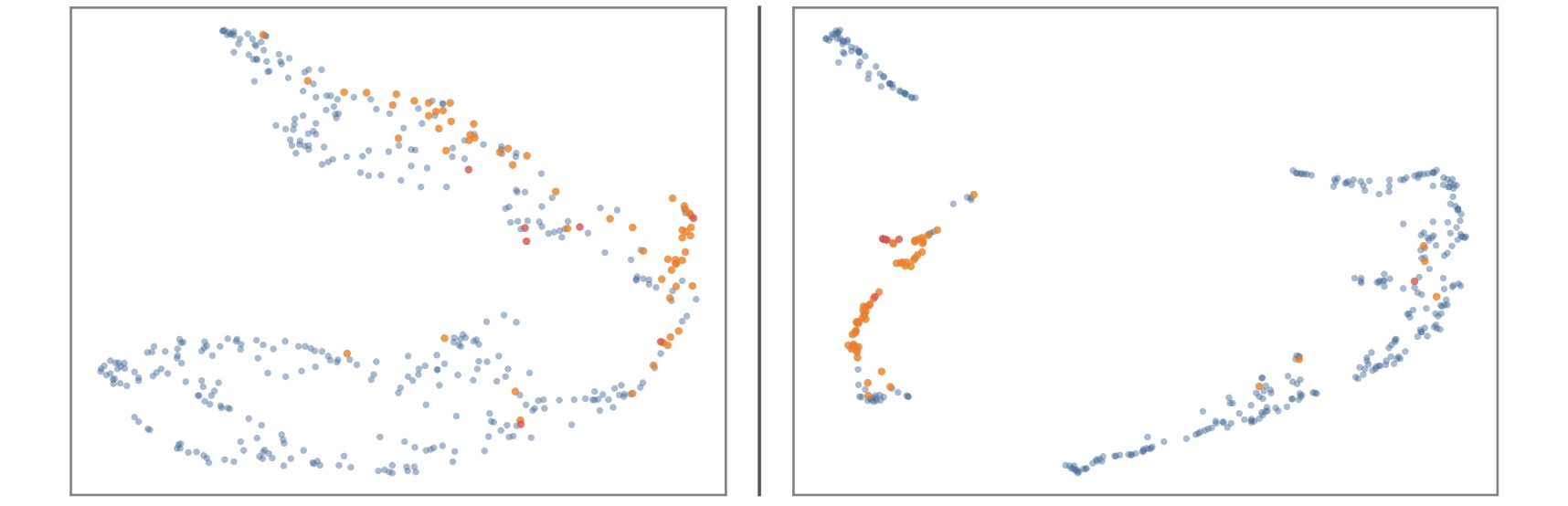}
        \caption{\emph{sheep}}
    
    \end{subfigure}

    \vspace{0.5em}

    \begin{subfigure}{0.95\textwidth}
        \centering
        \includegraphics[width=\textwidth]{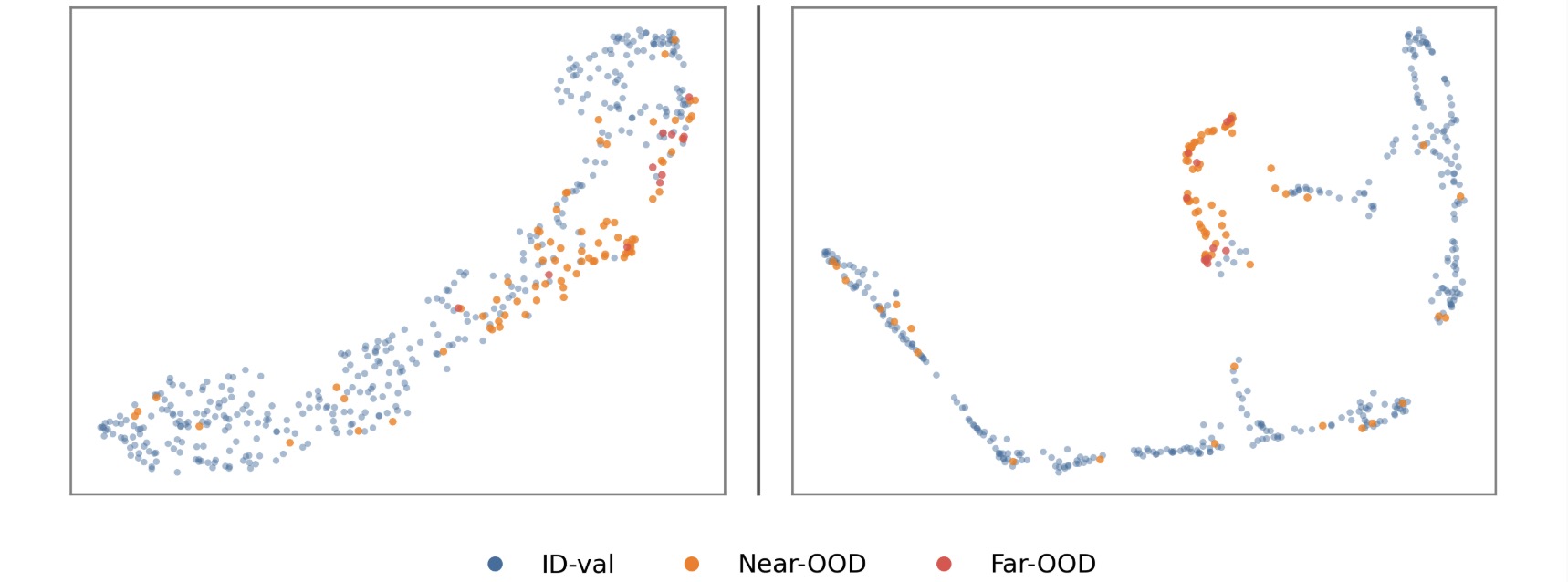}
        \caption{\emph{cat}}
        \label{fig:umap_cat}
    \end{subfigure}

    \caption{
        \textbf{UMAP visualization of representation spaces for detections from the \emph{dog}, \emph{sheep}, and \emph{cat} categories.} The visualizations are obtained using YOLO trained on PASCAL-VOC. We compare detector classification logits (left) with the proposed SPK representations (right), using the same ID-validation, Near-OoD, and Far-OoD samples. SPK yields a more structured representation space with clearer distributional differences between ID and OoD samples.
    }
    \label{fig:representation_space_umap}
\end{figure*}

\refstepcounter{subsection}
\subsection*{\thesubsection\quad Comprehensive Results on the Calibrated Benchmark}\label{app:moreResults}

\subsubsection{Effectiveness of SPK under AUROC}  Table~\ref{table:OoDAUROC} presents the OoD detection results measured by AUROC. Consistent with the FPR95 results reported in the main paper, these results further validate that SPK provides a more effective representation space for OoD detection when paired with the same OoD detection methods. The improved OoD detection performance of SPK is further supported by the enhanced separability observed in the learned representation space. Fig.~\ref{fig:representation_space_umap} visualizes the representation spaces of the object detector classification logits and SPK representations for detections predicted as \emph{dog}, \emph{sheep}, and \emph{cat} by a YOLO detector trained on PASCAL-VOC. While ID and OoD samples largely overlap in the conventional classification-logit space commonly used for OoD detection, SPK yields a more structured representation space with clearer ID--OoD separation.

\begin{table*}[h]
\centering
\caption{\textbf{AUROC comparison across detector architectures on PASCAL-VOC and BDD-100K.}
Higher is better.}\label{table:OoDAUROC}

\setlength{\tabcolsep}{3.2pt}
\renewcommand{\arraystretch}{1.08}

\resizebox{\textwidth}{!}{
\begin{tabular}{lcccc|cccc|cccc}
\toprule

\multirow{3}{*}{\textbf{Method}}
& \multicolumn{4}{c|}{\textbf{YOLO}}
& \multicolumn{4}{c|}{\textbf{Faster R-CNN}}
& \multicolumn{4}{c}{\textbf{RT-DETR}} \\

\cmidrule(lr){2-5}
\cmidrule(lr){6-9}
\cmidrule(lr){10-13}

& \multicolumn{2}{c}{\textbf{PASCAL-VOC}}
& \multicolumn{2}{c|}{\textbf{BDD-100K}}
& \multicolumn{2}{c}{\textbf{PASCAL-VOC}}
& \multicolumn{2}{c|}{\textbf{BDD-100K}}
& \multicolumn{2}{c}{\textbf{PASCAL-VOC}}
& \multicolumn{2}{c}{\textbf{BDD-100K}} \\

\cmidrule(lr){2-3}
\cmidrule(lr){4-5}
\cmidrule(lr){6-7}
\cmidrule(lr){8-9}
\cmidrule(lr){10-11}
\cmidrule(lr){12-13}

& \textbf{Near-OoD}
& \textbf{Far-OoD}
& \textbf{Near-OoD}
& \textbf{Far-OoD}
& \textbf{Near-OoD}
& \textbf{Far-OoD}
& \textbf{Near-OoD}
& \textbf{Far-OoD}
& \textbf{Near-OoD}
& \textbf{Far-OoD}
& \textbf{Near-OoD}
& \textbf{Far-OoD} \\

\midrule

MSP
& 81.24 & 79.47
& 77.63 & 75.12
& 78.71 & 73.84
& 72.48 & 76.39
& 79.16 & 78.72
& 74.31 & 75.06 \\

EBO
& 60.73 & 57.92
& 65.41 & 62.86
& 82.58 & 86.14
& 52.07 & 50.83
& 40.12 & 43.31
& 49.68 & 52.74 \\

MLS
& 59.41 & 61.26
& 63.78 & 65.02
& 84.76 & 83.58
& 54.82 & 59.46
& 56.37 & 59.74
& 54.91 & 57.43 \\

SCALE
& 69.84 & 71.73
& 72.46 & 79.31
& 57.49 & 69.72
& 64.18 & 67.82
& 69.37 & 75.16
& 66.42 & 65.73 \\

MDS
& 85.62 & 77.81
& 80.34 & 68.46
& 88.39 & 84.41
& 73.68 & 70.94
& 87.43 & 85.71
& 73.27 & 71.18 \\

BAM
& 90.14 & 89.38
& 89.57 & 86.94
& 80.96 & 91.56
& 80.75 & 83.65
& 75.75 & 81.65
& 76.50 & 80.45 \\

KNN
& 89.52 & 90.31
& 91.26 & 88.47
& 81.73 & 92.58
& 88.43 & 86.72
& 75.12 & 81.46
& 83.68 & 85.29 \\

iForest
& 77.38 & 80.71
& 82.46 & 80.19
& 76.31 & 85.64
& 81.37 & 83.29
& 70.82 & 81.94
& 78.46 & 83.57 \\

\midrule

\rowcolor[gray]{0.9}
\textbf{SPK-MDS}
& 96.12 & 93.55
& 95.80 & 98.65
& 96.10 & 94.20
& 92.10 & 91.80
& 94.35 & 95.54
& 89.10 & 99.21 \\

\rowcolor[gray]{0.9}
\textbf{SPK-BAM}
& 94.55 & 92.50
& 96.90 & 99.35
& 96.45 & 95.40
& 98.30 & 98.85
& 93.75 & 93.90
& 93.45 & 99.18 \\

\rowcolor[gray]{0.9}
\textbf{SPK-KNN}
& 94.91 & 93.31
& 97.40 & 99.71
& 97.07 & 96.46
& 99.06 & 99.36
& 94.40 & 94.87
& 93.94 & 99.36 \\

\rowcolor[gray]{0.9}
\textbf{SPK-iForest}
& \textbf{96.31} & \textbf{95.43}
& \textbf{98.10} & \textbf{99.81}
& \textbf{97.38} & \textbf{97.25}
& \textbf{99.50} & \textbf{99.65}
& \textbf{95.23} & \textbf{95.67}
& \textbf{95.78} & \textbf{99.60} \\

\bottomrule
\end{tabular}
}
\end{table*}

\subsubsection{Qualitative Examples} 
\begin{figure*}[h]
    \centering
    \includegraphics[width=\textwidth]{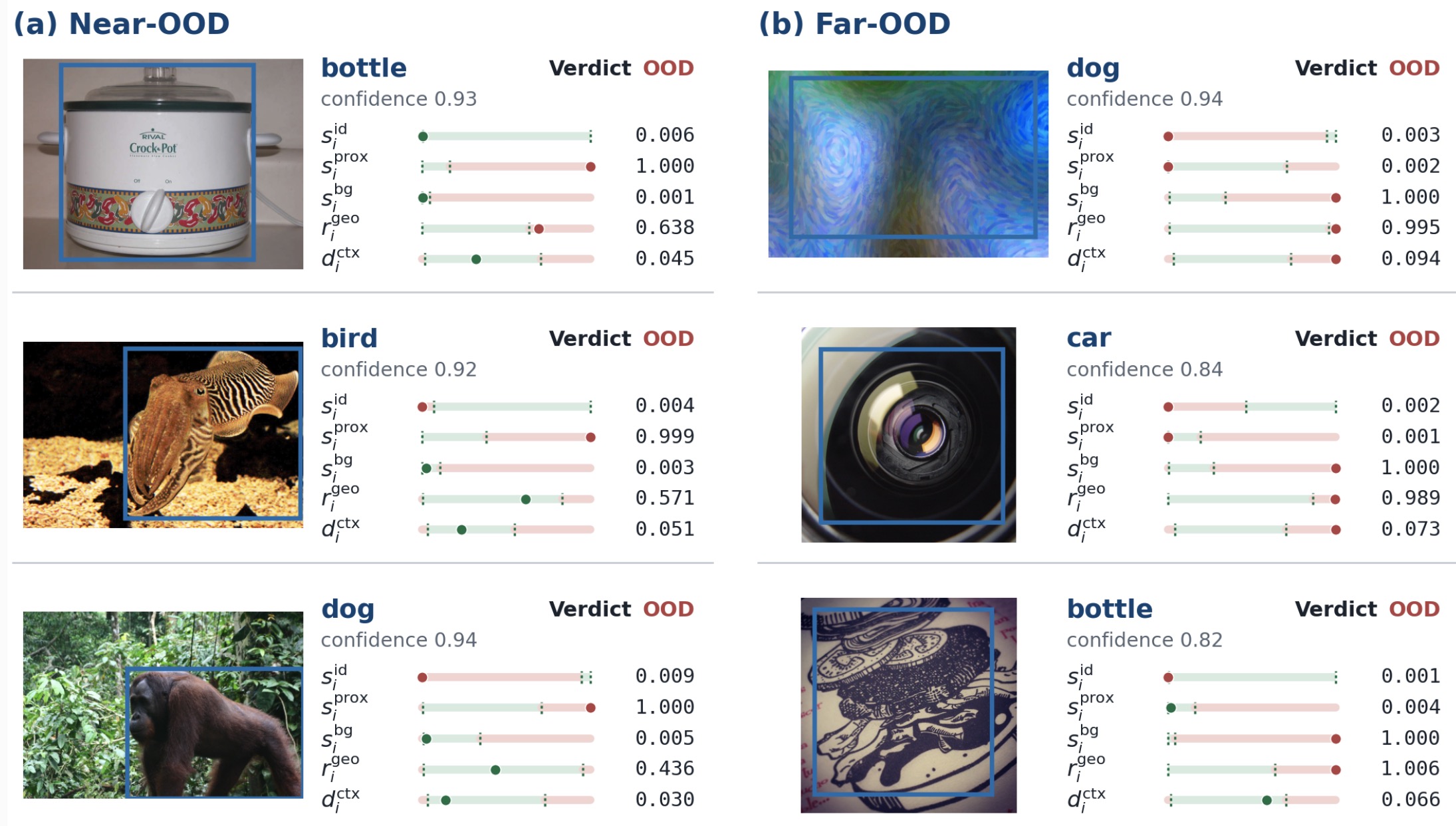}
    \caption{\textbf{Qualitative SPK OoD verdicts for OoD-induced hallucination.} Results are shown for YOLO on PASCAL-VOC. Near-OoD examples (left) exhibit strong proximal-concept responses, whereas Far-OoD examples (right) are dominated by background-concept responses. Each example shows the predicted bounding box, detector confidence, and five SPK prior components. Green bands represent class-specific normal ranges estimated from ID training data, red bands indicate values outside these ranges, and circular markers denote the observed component values. All examples are correctly rejected as OoD.}
    \label{fig:spk-full-qualitative-verdicts}
\end{figure*}

Fig.~\ref{fig:spk-full-qualitative-verdicts} provides qualitative examples illustrating how SPK characterizes different sources of detector hallucinations despite high-confidence predictions. For Near-OoD examples, the hallucinated detections exhibit strong proximal-concept responses, while ID and background concept responses remain suppressed, indicating that these false positives are primarily caused by their semantic similarity to known categories. In contrast, Far-OoD examples lack coherent object-level semantics and are dominated by strong background responses with negligible ID or proximal activations. The geometric and contextual priors provide additional complementary evidence by capturing object-scale and image-level inconsistencies. Based on the resulting SPK representations, the OoD detector successfully rejects these hallucinated detections despite their high confidence scores. These examples demonstrate that SPK not only detects OoD-induced hallucinations but also provides interpretable evidence about which prior components contribute to each rejection decision.

\subsubsection{Ablation on Semantic Prior Learning} Additional ablation results on Faster R-CNN and RT-DETR are presented in Table~\ref{tab:appendix_loss_ablation}, further demonstrating the effectiveness of jointly optimizing the three proposed loss components.

\begin{table*}[h]
\centering
\caption{\textbf{Ablation of SPK loss components.}
Evaluated across Faster R-CNN and RT-DETR on PASCAL-VOC and BDD-100K. Lower FPR95 is better.}
\label{tab:appendix_loss_ablation}

\setlength{\tabcolsep}{4pt}
\renewcommand{\arraystretch}{1.10}

\resizebox{0.95\textwidth}{!}{
\begin{tabular}{ccc|cccc|cccc}
\toprule

\multirow{3}{*}{$\mathcal{L}_{\mathrm{dice}}$}
& \multirow{3}{*}{$\mathcal{L}_{\mathrm{suppress}}$}
& \multirow{3}{*}{$\mathcal{L}_{\mathrm{group}}$}
& \multicolumn{4}{c|}{\textbf{Faster R-CNN}}
& \multicolumn{4}{c}{\textbf{RT-DETR}} \\

\cmidrule(lr){4-7}
\cmidrule(lr){8-11}

& &
& \multicolumn{2}{c}{\textbf{PASCAL-VOC}}
& \multicolumn{2}{c|}{\textbf{BDD-100K}}
& \multicolumn{2}{c}{\textbf{PASCAL-VOC}}
& \multicolumn{2}{c}{\textbf{BDD-100K}} \\

\cmidrule(lr){4-5}
\cmidrule(lr){6-7}
\cmidrule(lr){8-9}
\cmidrule(lr){10-11}

& &
& \textbf{Near-OoD}
& \textbf{Far-OoD}
& \textbf{Near-OoD}
& \textbf{Far-OoD}
& \textbf{Near-OoD}
& \textbf{Far-OoD}
& \textbf{Near-OoD}
& \textbf{Far-OoD} \\

\midrule

\xmark & \cmark & \cmark
& 27.87
& 19.98
& 18.36
& 18.74
& 32.31
& 27.08
& 25.35
& 29.14 \\

\cmark & \cmark & \xmark
& 24.42
& 15.42
& 14.15
& 16.11
& 27.86
& 22.53
& 20.71
& 24.36 \\

\cmark & \xmark & \cmark
& 22.19
& 14.73
& 11.34
& 13.27
& 24.86
& 21.64
& 18.15
& 20.81 \\

\rowcolor[gray]{0.92}
\cmark & \cmark & \cmark
& \textbf{13.92}
& \textbf{10.52}
& \textbf{2.31}
& \textbf{1.52}
& \textbf{15.48}
& \textbf{17.32}
& \textbf{11.42}
& \textbf{9.25} \\

\bottomrule
\end{tabular}
}
\end{table*}

\subsubsection{Ablation on Prior Components.} Additional ablation results on Faster R-CNN and RT-DETR are presented in Table~\ref{tab:prior_component_ablation}, further confirming the effectiveness of jointly combining the three types of priors.

\begin{table*}[h]
\centering
\caption{\textbf{Ablation of SPK prior components.}
Evaluated across Faster R-CNN and RT-DETR on PASCAL-VOC and BDD-100K. Lower FPR95 is better.}
\label{tab:prior_component_ablation}

\setlength{\tabcolsep}{4pt}
\renewcommand{\arraystretch}{1.10}

\resizebox{0.95\textwidth}{!}{
\begin{tabular}{l|cccc|cccc}
\toprule

\textbf{Prior components}
& \multicolumn{4}{c|}{\textbf{Faster R-CNN}}
& \multicolumn{4}{c}{\textbf{RT-DETR}} \\

\cmidrule(lr){2-5}
\cmidrule(lr){6-9}

&
\multicolumn{2}{c}{\textbf{PASCAL-VOC}}
&
\multicolumn{2}{c|}{\textbf{BDD-100K}}
&
\multicolumn{2}{c}{\textbf{PASCAL-VOC}}
&
\multicolumn{2}{c}{\textbf{BDD-100K}} \\

\cmidrule(lr){2-3}
\cmidrule(lr){4-5}
\cmidrule(lr){6-7}
\cmidrule(lr){8-9}

&
\textbf{Near-OoD}
&
\textbf{Far-OoD}
&
\textbf{Near-OoD}
&
\textbf{Far-OoD}
&
\textbf{Near-OoD}
&
\textbf{Far-OoD}
&
\textbf{Near-OoD}
&
\textbf{Far-OoD} \\

\midrule

{\scriptsize\bfseries Semantic}
& 18.17
& 18.91
& 23.92
& 25.21
& 18.34
& 29.17
& 33.36
& 34.75 \\

{\scriptsize\bfseries Semantic + Geometric}
& 15.78
& 16.43
& 14.52
& 15.71
& 15.85
& 25.35
& 26.51
& 26.82 \\

\rowcolor[gray]{0.92}
{\scriptsize\bfseries Semantic + Geometric + Contextual}
& \textbf{13.92}
& \textbf{10.52}
& \textbf{2.31}
& \textbf{1.52}
& \textbf{15.48}
& \textbf{17.32}
& \textbf{11.42}
& \textbf{9.25} \\

\bottomrule
\end{tabular}
}
\end{table*}

\refstepcounter{subsection}
\subsection*{\thesubsection\quad Experimental Results on Uncalibrated Benchmark}\label{app:uncalibratedBenchmark}

To facilitate comparison with a broader range of existing OoD detection methods, we additionally evaluate SPK under the conventional uncalibrated benchmark adopted by prior work. While our main experiments follow the calibrated benchmark proposed in~\cite{wu2026revisiting}, several previous methods are architecture-specific, require specialized training protocols, or lack publicly available implementations, making direct evaluation under the calibrated benchmark difficult. Evaluating under their original protocol therefore enables a broader and fairer comparison with the existing literature, despite the known annotation issues of this benchmark discussed in~\cite{wu2026revisiting}.

Following~\cite{Peng_2026_CVPR}, we evaluate Deformable-DETR on PASCAL-VOC and BDD-100K, and Faster R-CNN on PASCAL-VOC, with MS-COCO~\cite{lin2014microsoft} and OpenImages~\cite{kuznetsova2020open} as OoD test sets. Tables~\ref{tab:ood_comparison_deformable_detr} and~\ref{tab:ood_od_fasterrcnn} show that SPK variants consistently achieve state-of-the-art or highly competitive performance across all evaluation settings. Among the twelve evaluation cases, SPK variants achieves the best performance in ten, while ranking second in the remaining case. Some existing methods additionally exploit external visual encoders for OoD detection rather than relying solely on detector-intrinsic representations. To enable a fair comparison under this setting, we introduce SPK (DINO ViT), which replaces the detector-intrinsic contextual prior with an image-level representation extracted by DINO ViT. This variant follows the same external-encoder setting as methods such as UNO-Adapter while retaining the semantic and geometric priors of SPK.

\begin{table*}[t]
\centering
\caption{
\textbf{Comparison with competitive OoD detection methods for Deformable-DETR.} Results are reported on PASCAL-VOC and BDD-100K as ID datasets, with MS-COCO and OpenImages as OoD datasets. Higher AUROC and lower FPR95 indicate better OoD detection performance. Methods marked with $\dagger$ employ an external DINO ViT encoder to extract additional visual representations for OoD detection, rather than relying solely on Deformable-DETR features. SPK (DINO ViT) is included to provide a fair comparison with UNO-Adapter, as both methods operate under this setting. The best results are highlighted in bold.}
\label{tab:ood_comparison_deformable_detr}

\setlength{\tabcolsep}{4.0pt}
\renewcommand{\arraystretch}{1.08}

\resizebox{\textwidth}{!}{
\begin{tabular}{lcccc|cccc}
\toprule

\multirow{3}{*}{\textbf{Method}}
& \multicolumn{4}{c|}{\textbf{ID: PASCAL-VOC}}
& \multicolumn{4}{c}{\textbf{ID: BDD-100K}} \\

\cmidrule(lr){2-5}
\cmidrule(lr){6-9}

& \multicolumn{2}{c}{\textbf{OoD: MS-COCO}}
& \multicolumn{2}{c|}{\textbf{OoD: OpenImages}}
& \multicolumn{2}{c}{\textbf{OoD: MS-COCO}}
& \multicolumn{2}{c}{\textbf{OoD: OpenImages}} \\

\cmidrule(lr){2-3}
\cmidrule(lr){4-5}
\cmidrule(lr){6-7}
\cmidrule(lr){8-9}

& \textbf{FPR95$\downarrow$}
& \textbf{AUROC$\uparrow$}
& \textbf{FPR95$\downarrow$}
& \textbf{AUROC$\uparrow$}
& \textbf{FPR95$\downarrow$}
& \textbf{AUROC$\uparrow$}
& \textbf{FPR95$\downarrow$}
& \textbf{AUROC$\uparrow$} \\

\midrule

MDS~\cite{lee2018simple}
& 97.39 & 50.28
& 97.88 & 49.08
& 70.86 & 76.83
& 71.43 & 77.98 \\

Gram matrices~\cite{sastry2020detecting}
& 94.16 & 43.97
& 95.29 & 38.81
& 73.81 & 60.13
& 71.56 & 57.14 \\

KNN~\cite{sun2022out}
& 91.80 & 62.15
& 91.36 & 59.64
& 64.75 & 80.90
& 61.13 & 79.64 \\

CSI~\cite{tack2020csi}
& 84.00 & 55.07
& 79.16 & 51.37
& 70.27 & 77.93
& 71.30 & 76.42 \\

VOS~\cite{du2022towards}
& 97.46 & 54.40
& 97.07 & 52.77
& 76.44 & 77.33
& 72.58 & 76.62 \\

OW-DETR~\cite{gupta2022ow}
& 93.09 & 55.70
& 93.82 & 57.80
& 80.78 & 70.29
& 77.37 & 73.78 \\

DisMax~\cite{macedo2022distinction}
& 82.05 & 75.21
& 76.37 & 70.66
& 77.62 & 72.14
& 81.23 & 67.18 \\

SIREN-vMF~\cite{du2022siren}
& 75.49 & 76.10
& 78.36 & 71.05
& 67.54 & 80.06
& 66.31 & 79.77 \\

SIREN-KNN~\cite{du2022siren}
& 64.77 & 78.23
& 65.99 & 74.93
& 53.97 & 86.56
& 47.28 & 89.00 \\

SAFE~\cite{wilson2023safe}
& 48.88 & 78.88
& \textbf{8.99} & \textbf{96.73}
& 39.18 & 85.95
& 21.10 & 94.31 \\

InfoBound~\cite{zhu2025infobound}
& 44.88 & 89.76
& 43.89 & 88.00
& 44.88 & 89.76
& 43.89 & 88.00 \\

UNO-Adapter$^{\dagger}$~\cite{Peng_2026_CVPR}
& 32.61 & 91.68
& 19.90 & 95.40
& 9.88 & 97.61
& 3.80 & 99.04 \\

\midrule

\rowcolor[gray]{0.9}
\textbf{SPK}
& 52.32 & 75.84
& 24.38 & 90.20
& 1.68 & 99.42
& 0.37 & 99.93 \\

\rowcolor[gray]{0.9}
\textbf{SPK (DINO ViT)$^{\dagger}$}
& \textbf{28.55} & \textbf{92.38}
& 14.85 & 96.25
& \textbf{0.00} & \textbf{99.80}
& \textbf{0.00} & \textbf{99.97} \\

\bottomrule
\end{tabular}
}
\end{table*}

\begin{table*}[h]
\centering
\caption{\textbf{Comparison with competitive OoD detection methods for Faster R-CNN.} Results are reported on PASCAL-VOC as the ID dataset, with MS COCO and OpenImages as OoD datasets. Higher AUROC and lower FPR95 indicate better OoD detection performance. Methods marked with $\dagger$ employ an external DINO ViT encoder to extract additional visual representations for OoD detection, rather than relying solely on Faster R-CNN features. SPK (DINO ViT) is included to provide a fair comparison with UNO-Adapter, as both methods operate under this setting. The best results are highlighted in bold.}
\label{tab:ood_od_fasterrcnn}
\resizebox{0.68\linewidth}{!}{
\begin{tabular}{l|cc|cc}
\hline
\multirow{2}{*}{Method}
& \multicolumn{2}{c|}{MS-COCO}
& \multicolumn{2}{c}{OpenImages} \\
& AUROC$\uparrow$ & FPR95$\downarrow$
& AUROC$\uparrow$ & FPR95$\downarrow$ \\
\hline
CSI~\cite{tack2020csi}             & 82.95 & 57.41 & 81.83 & 59.91 \\
GAN-Synthesis~\cite{lee2018training}   & 82.67 & 59.97 & 83.67 & 60.93 \\
VOS~\cite{du2022towards}             & 85.23 & 51.33 & 88.70 & 47.53 \\
SIREN~\cite{du2022siren}           & 85.36 & 64.68 & 82.78 & 68.53 \\
TIB~\cite{wu2023tib}             & 90.36 & 41.55 & 88.09 & 47.19 \\
DFDD~\cite{wu2023deep}            & 90.79 & 41.34 & 88.65 & 44.52 \\
WFS~\cite{wu2025percept}             & 89.01 & 40.05 & 90.35 & 39.17 \\
UNO-Adapter$^{\dagger}$~\cite{Peng_2026_CVPR}     & 91.25 & 38.73 & 92.40 & 35.74 \\
\rowcolor[gray]{0.9} 
\textbf{SPK} & 91.58 & 41.20 & 95.50 & 25.61 \\
\rowcolor[gray]{0.9} 
\textbf{SPK (DINO ViT)$^{\dagger}$} & \textbf{95.48} & \textbf{26.28} & \textbf{98.16} & \textbf{10.97} \\
\hline
\end{tabular}
}
\end{table*}

\refstepcounter{subsection}
\subsection*{\thesubsection\quad Inference Runtime Analysis.}\label{app:overhead}

Table~\ref{tab:computational_cost} reports the per-image runtime of SPK and its individual components for a YOLO model pretrained on PASCAL-VOC. The original detector inference takes $10.15$ ms per image on an NVIDIA A4000-8GB GPU. By extracting detection outputs, image-level contextual embeddings, and RoI features within the same forward pass, SPK inference takes $12.65$ ms.  Semantic prior elicitation and Isolation Forest inference introduce only marginal additional latency. Although detector inference and RoI feature extraction costs may vary across object detector architectures, the remaining components have comparable computational costs across detectors, as they operate on compact representations with lightweight additional modules.
\begin{table*}[h]
\centering
\small
\caption{\textbf{Per-image runtime of the complete SPK inference pipeline.}  Reported for a YOLO model pretrained on PASCAL-VOC. The complete SPK pipeline introduces an additional $2.72$ ms latency per image, corresponding to a $26.8\%$ runtime overhead over the original detector inference. \textsuperscript{*}SPK inference obtains detection outputs, image-level contextual embeddings, and RoI features within the same forward pass.}
\label{tab:computational_cost}
\setlength{\tabcolsep}{6pt}
\renewcommand{\arraystretch}{1.08}

\resizebox{0.32\linewidth}{!}{
\begin{tabular}{@{}lc@{}}
\toprule
\textbf{Component} & \textbf{Cost (ms)} \\
\midrule
Original inference 
    & 10.15 \\
SPK inference\textsuperscript{*}
    & 12.65 \\
Semantic prior elicitation
    & 0.17 \\
Isolation Forest
    & 0.05 \\
\bottomrule
\end{tabular}
}
\end{table*}

\end{document}